\documentclass[10pt,twocolumn,letterpaper]{article}

\usepackage[pagenumbers]{cvpr} 

\usepackage{graphicx}
\usepackage{amsmath}
\usepackage{amssymb}
\usepackage{booktabs}

\usepackage{mathtools}
\DeclarePairedDelimiter\norm{\lVert}{\rVert}%
\usepackage{bm}
\usepackage{subcaption}
\usepackage{algpseudocode}
\usepackage[linesnumbered,ruled]{algorithm2e}
\usepackage{booktabs}
\usepackage[dvipsnames]{xcolor}
\usepackage{nccmath}
\usepackage{array}
\usepackage{enumitem}
\usepackage{siunitx}
\usepackage{multirow}
\newcolumntype{P}[1]{>{\centering\arraybackslash}p{#1}}
\newcolumntype{M}[1]{>{\centering\arraybackslash}m{#1}}
\usepackage{pifont}
\newcommand{\comment}[1]{}

\usepackage[pagebackref,breaklinks,colorlinks]{hyperref}

\usepackage[capitalize]{cleveref}
\crefname{section}{Sec.}{Secs.}
\Crefname{section}{Section}{Sections}
\Crefname{table}{Table}{Tables}
\crefname{table}{Tab.}{Tabs.}

\def\cvprPaperID{*****} 
\def\confName{CVPR}
\def\confYear{2022}

\begin{document}

\title{Ego2HandsPose: A Dataset for Egocentric Two-hand 3D Global Pose Estimation}

\author{Fanqing Lin\\
Brigham Young University\\
{\tt\small fanqinglin@byu.edu}
\and
Tony Martinez\\
Brigham Young University\\
{\tt\small martinez@cs.byu.edu}
}
\maketitle

\begin{abstract}
Color-based two-hand 3D pose estimation in the global coordinate system is essential in many applications. However, there are very few datasets dedicated to this task and no existing dataset supports estimation in a non-laboratory environment. This is largely attributed to the sophisticated data collection process required for 3D hand pose annotations, which also leads to difficulty in obtaining instances with the level of visual diversity needed for estimation in the wild. Progressing towards this goal, a large-scale dataset Ego2Hands was recently proposed to address the task of two-hand segmentation and detection in the wild. The proposed composition-based data generation technique can create two-hand instances with quality, quantity and diversity that generalize well to unseen domains. In this work, we present Ego2HandsPose, an extension of Ego2Hands that contains 3D hand pose annotation and is the first dataset that enables color-based two-hand 3D tracking in unseen domains. To this end, we develop a set of parametric fitting algorithms to enable 1) 3D hand pose annotation using a single image, 2) automatic conversion from 2D to 3D hand poses and 3) accurate two-hand tracking with temporal consistency. We provide incremental quantitative analysis on the multi-stage pipeline and show that training on our dataset achieves state-of-the-art results that significantly outperforms other datasets for the task of egocentric two-hand global 3D pose estimation.
\end{abstract}
\section{Introduction}
Hand pose estimation and tracking is significant for many applications that involve Human-Computer Interaction (HCI), gesture recognition and sign language recognition. Particularly, as VR/AR/MR applications gain rapid development with the trending of metaverse, two-hand tracking is becoming a fundamental feature for an immersive user experience. In addition, due to the overhead cost of multi-camera setups as well as depth cameras, there is motivation to approach this task using a single ubiquitous RGB camera. However, there is limited attention from the community on color-based two-hand application, which is an extremely challenging task requiring steps not needed in single-hand tasks that use cropped images as input: two-hand detection/segmentation, robustness against inter-hand occlusion as well as 3D global hand pose estimation with accurate absolute 3D joint positions.\\
\begin{figure}
  \begin{subfigure}[b]{\linewidth}
  \centering
  \includegraphics[width=0.9\linewidth]{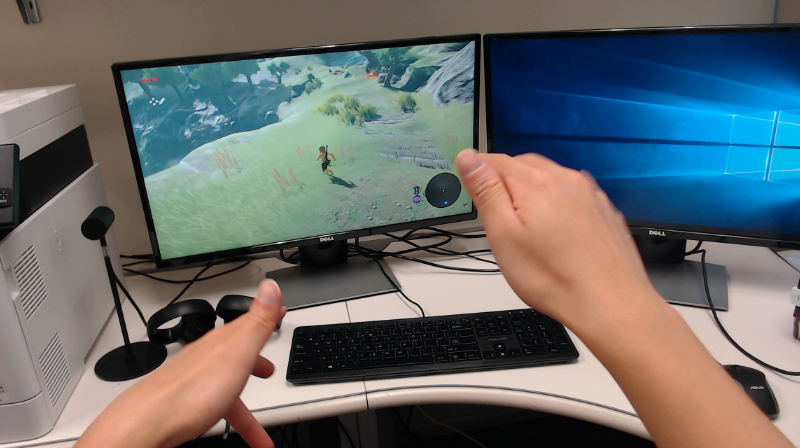}
  \end{subfigure}
  \begin{subfigure}[b]{\linewidth}
  \centering
  \vspace{0.1cm}
  \includegraphics[width=0.9\linewidth]{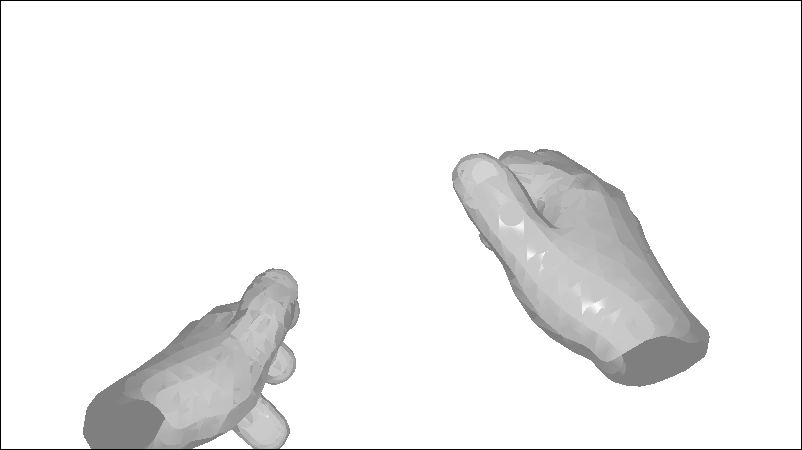}
  \end{subfigure}
  \caption{Our method enables 3D hand pose annotation using a single image. We show a sample image from the test set of Ego2Hands (top) and visualization of its corresponding two-hand pose annotation (bottom).}
  \label{fig:intro_img}
  \vspace{-6mm}
\end{figure}
\indent Existing color-based two-hand pose datasets~\cite{Moon20, Brahmbhatt20} are captured in third-person viewpoints with multiple cameras and laboratory backgrounds. Two-hand datasets~\cite{Wang20, Tzionas16} with RGB-D data can remove background using depth thresholding or a green screen. However, these data have limited accuracy from depth-based tracking and lack visual diversity due to the small number of participants. As a result, methods trained on existing datasets cannot generalize to the real-world domain. In addition to limited diversity, although data for third-person viewpoint has many applications, it constrains the users to perform gestures in front of a fixed camera. On the other hand, hand tracking in egocentric viewpoint has no such constraint and has increasing demand in VR/AR applications. To the best of our knowledge, there is currently no real-world RGB dataset for egocentric two-hand tracking in the wild.\\
\indent To enable two-hand applications using a single RGB camera in a domain invariant setting,~\cite{Lin20} first proposed a large-scale two-hand segmentation/detection dataset named Ego2Hands. Unlike traditional hand segmentation datasets~\cite{Bambach15, Khan18, Li18} with limited quantity/diversity and generalization ability, Ego2Hands composites two-hand instances at training time with excellent generalization to unseen environments. However, Ego2Hands does not contain hand pose annotation. In this work, we develop a parametric fitting algorithm ManoFit that can fit the deformable MANO hand model~\cite{Romero17} given an arbitrary number of 2D joint locations and minimal manual guidance towards the optimal solution. This is the first tool that enables users to annotate 3D hand poses using a single image, which significantly simplifies the annotation process compared to previous methods (see Section \ref{sec:related_works}) and provides open access for the community to generate additional hand pose data. Using our method, we create a new dataset by manually annotating$\sim7,000$ selected frames with diversity from the training set and$\sim2,000$ frames from the test set of Ego2Hands (see Figure \ref{fig:intro_img}). We introduce Ego2HandsPose, the first dataset that enables egocentric two-hand 3D global pose estimation in the wild using a single camera.\\
\indent In addition to Ego2handsPose, as there are existing datasets with only 2D hand pose annotations, we apply ManoFit to automatically convert HIU-DMTL~\cite{Zhang21}, PanHand2D~\cite{Simon17} and OneHand10k~\cite{Wang18} to 3D hand pose datasets. Manual validation is performed on all generated instances to remove dirty data. For quantitative analysis on the accuracy of our generated hand poses, we evaluate ManoFit on FreiHAND~\cite{Zimmermann19} which contains both 2D and 3D hand pose annotation and show convincing results.\\
\indent To validate Ego2HandsPose for the task of two-hand 3D tracking, we follow \cite{Lin19} and use a multi-stage pipeline to estimate the segmentation/detection and 2D/3D canonical hand pose for both hands. As the 3D canonical hand pose estimation network uses input in the form of heatmaps, there is no need for the training to be constrained by 3D hand poses of actual images and we take a novel approach by training on generated poses. We introduce this synthetic dataset as MANO3Dhands, which generates 3D hand poses based on our collected real-world pose distribution. Cross-dataset evaluation shows that MANO3Dhands has the best generalization score compared to existing 3D hand pose datasets. For the last stage of the pipeline, we modify ManoFit to achieve temporally consistent two-hand 3D tracking by using the estimated 2D and 3D canonical joint locations. Quantitative analysis shows that training on Ego2HandsPose achieves over $30\%$ improvement compared to the second top dataset for the task of two-hand 2D, 3D canonical and 3D global pose estimation.
\section{Related Work}\label{sec:related_works}
In this section, we introduce relevant RGB-based 3D hand pose datasets for single-hand and two-hand scenarios. \\
\noindent\textbf{Single-hand 3D pose datasets.} Obtaining 3D hand pose annotation on real-world images is a challenging problem that commonly requires extensive manual annotation using multi-view setups or RGB-D data to resolve the depth ambiguity and self-occlusion. Early work~\cite{Zhang16} proposed the \textit{Stereo Tracking Benchmark} (\textit{STB}) dataset that includes a single participant performing simple gestures with 6 backgrounds. To generate data with quantity and diversity, pioneering work~\cite{Zimmermann17} introduced the \textit{Rendered Hand Pose Dataset} (\textit{RHD}) using 20 rendered characters performing 39 static gestures, which supported the training and evaluation of a CNN-based two-stage pipeline that showed 3D hand pose estimation is achievable using a single image. In an attempt to bridge the domain gap between synthetic and real-world data, \textit{GANeratedHands}~\cite{Mueller18} (\textit{GANHands}) was proposed using a CycleGAN~\cite{Zhu17} approach. Using the CMU Panoptic Studio with 10 RGB-D sensors, 480 VGA and 31 HD cameras,~\cite{Joo18} proposed the \textit{Panoptic Hand} (\textit{PanHand3D}) dataset labeled using multiview Bootstrapping~\cite{Simon17}. To achieve better cross-dataset generalization,~\cite{Zimmermann19} proposed \textit{FreiHAND} that was captured using 8 calibrated \& synchronized cameras in a green screen setting, which allowed for additional diversity from background replacement.\\
\noindent\textbf{Two-hand 3D pose datasets.} Inter-hand occlusion and interaction can introduce additional challenges for 3D hand pose annotation. To address two-hand pose estimation with object handling,~\cite{Tzionas16} proposed the \textit{Tzionas Dataset} collected using RGB-D data and a combination of a generative linear blend skinning model~\cite{Lewis00} and a discriminative model trained on manually annotated finger tips. To provide more data with two-hand interaction,~\cite{Wang20} proposed \textit{RGB2Hands} that was captured using a RGB-D sensor and labeled with a depth-based two hand tracker~\cite{Mueller19}. However, the obtained annotation can be erroneous and synthetic data was used to complement this issue. Focusing on two-hand object grasping,~\cite{Brahmbhatt20} proposed \textit{ContactPose} that was captured using 7 calibrated RGB cameras, 3 RGB-D cameras and one thermal camera for contact capture. 3D hand Pose annotation was obtained using estimated 2D keypoints as well as extracted object pose and contact locations. Taking a different direction, ~\cite{Moon20} focused on two-hand close interaction without objects and proposed \textit{InterHand2.6M}. The data was collected in a multi-view studio consisting of 80-140 cameras and annotated in a two-stage pipeline. The first stage consists of extensive manual annotation of 2D hand poses followed by 3D triangulation. The second stage utilizes an automatic 2D annotator trained from data in the first stage and triangulation for 3D keypoints. Recognizing the importance of egocentric data, \textit{H2O}~\cite{Kwon21} was captured with 5 RGB-D cameras with 1 mounted on the helmet in 3 environments. Two-hand poses with object manipulation annotation was obtained by fitting the MANO model to multi-view depth-data and estimated 2D poses from OpenPose~\cite{Cao17}.
\begin{figure}[t]
  \small
  \centering
  \begin{subfigure}[t]{0.45\linewidth}
    \includegraphics[width=\linewidth]{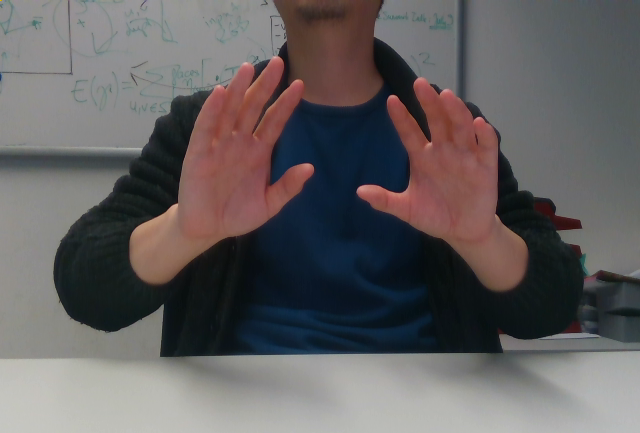}
    \caption{RGB2Hands}
  \end{subfigure}
  \begin{subfigure}[t]{0.45\linewidth}
    \includegraphics[width=\linewidth]{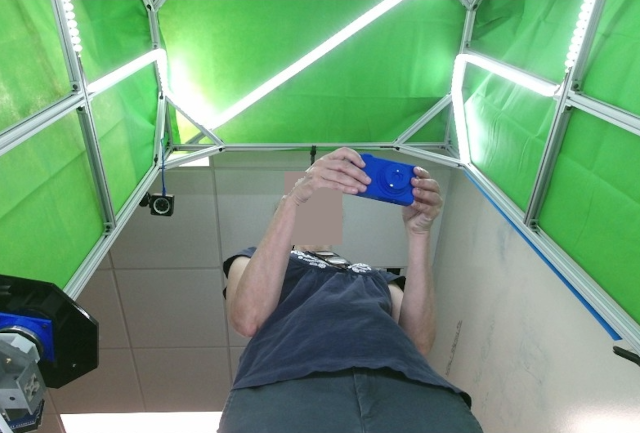}
    \caption{ContactPose}
  \end{subfigure}
  \begin{subfigure}[t]{0.45\linewidth}
    \includegraphics[width=\linewidth]{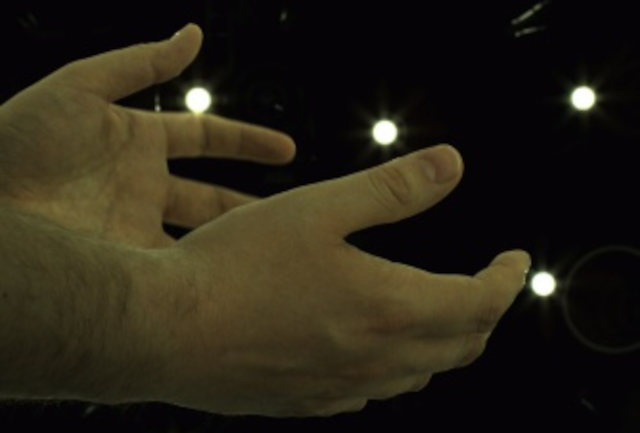}
    \caption{InterHand2.6M}
  \end{subfigure}
  \begin{subfigure}[t]{0.45\linewidth}
    \includegraphics[width=\linewidth]{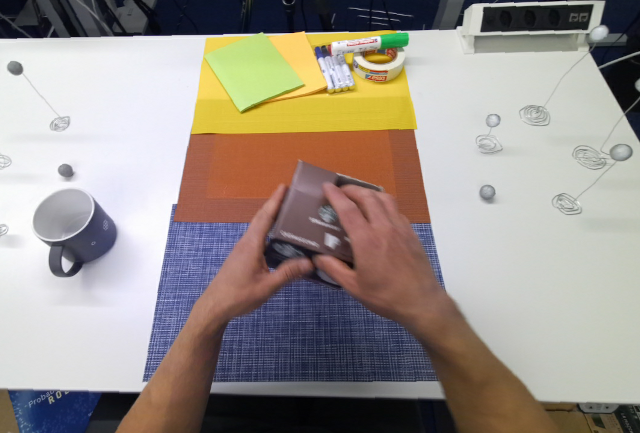}
    \caption{H2O}
  \end{subfigure}
  \vspace{-2mm}
  \caption{Sample images from existing two-hand pose datasets.}
  \label{fig:dataset_samples}
  \vspace{-4mm}
\end{figure}
\section{Ego2HandsPose}
\indent There are two major motivations for the introduction of Ego2HandsPose. First, Figure \ref{fig:dataset_samples} shows that existing datasets for RGB-based two-hand pose estimation all contain images captured in laboratory environments with limited diversity. Consequently, models trained on these datasets cannot generalize to other unseen environments or be applied to practical applications. Second, there is limited RGB data that address two-hand pose estimation in the egocentric viewpoint, which does not constrain the user to be in front of the fixed camera and is essential in applications such as VR/AR/MR. As the only existing RGB-based egocentric two-hand pose dataset, \textbf{H2O} has a heavy emphasis on manipulation of 8 objects in 3 scenes with 4 participants, which results in limited pose space and visual diversity for the environment and the target hands.\\ 
\indent For two-hand 3D tracking, segmentation and detection of both hands are commonly required prior to pose estimation~\cite{Taylor17, Mueller19, Lin19}. Recently,~\cite{Lin20} introduced Ego2Hands for the task of two-hand segmentation and detection in the wild. It consists of a training set captured in a green screen setting with$\sim180k$ right hand instances from 22 participants and composites two-hand images at training time by horizontally flipping one right hand to create the left hand. This approach circumvents the issue of data scarcity for two-hand segmentation/detection and allows for sufficient diversity necessary for estimation in the wild. For evaluation, Ego2Hands provides a test set consisting of 8 sequences collected with diverse scenes, lighting and skin tones. Despite enabling models to achieve a promising level of generalization on two-hand segmentation and detection, Ego2Hands does not provide any hand pose annotation.\\
\indent Existing 3D hand pose annotation methods require either RGB-D or calibrated multi-view RGB setups for data collection. Extensive multi-view manual annotation on 2D keypoints as well as triangulation are commonly required to subsequently extract the 3D hand poses. We argue that the resources required by existing annotation methods are not commonly available in the community, which significantly limits the quantity and diversity of hand pose data available in general and consequently impacts hand pose related research potential. To address this issue, we introduce an annotation tool that utilizes a parametric fitting algorithm \textit{ManoFit} with manual guidance to enable 3D hand pose annotation using a single RGB image.
\begin{figure}[t]
  \small
  \centering
  \begin{subfigure}[t]{0.32\linewidth}
    \includegraphics[width=\linewidth]{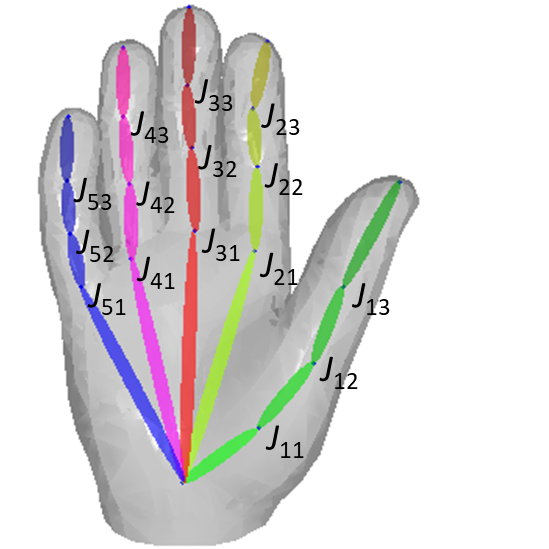}
  \end{subfigure}
  \begin{subfigure}[t]{0.32\linewidth}
    \includegraphics[width=\linewidth]{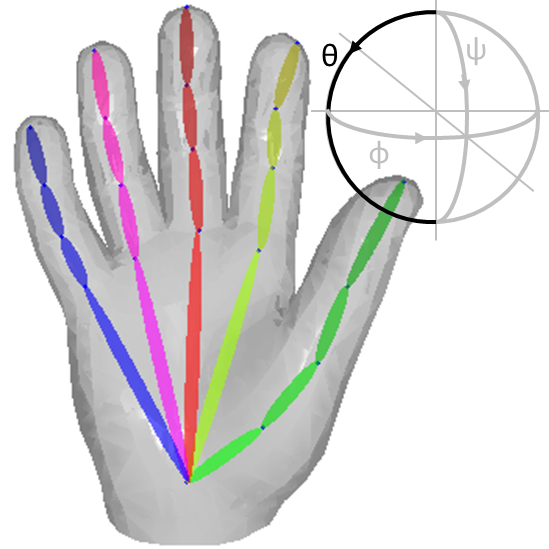}
  \end{subfigure}
  \begin{subfigure}[t]{0.32\linewidth}
    \includegraphics[width=\linewidth]{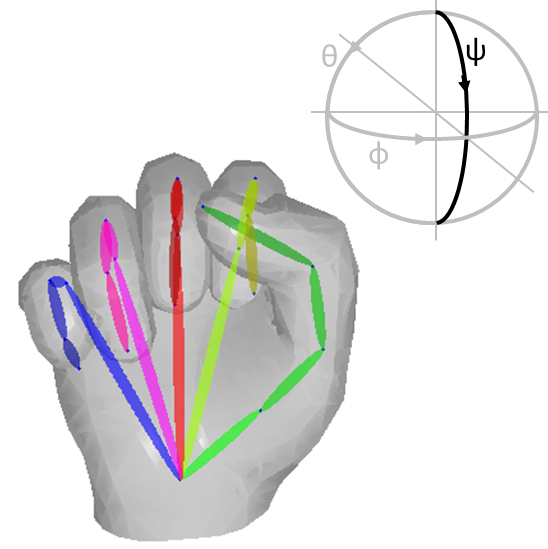}
  \end{subfigure}
  \caption{MANO hand representation for joints and axis angles. The first image shows a default hand pose. The second and third image show poses with joints rotated in the two primary directions.}
  \label{fig:mano_samples}
  \vspace{-4mm}
\end{figure}
\subsection{Supervised ManoFit}\label{sec:manofit1}
The differentiable MANO hand model proposed by~\cite{Romero17} is widely used for hand pose estimation and fitting. It is parameterized by $P \in \mathcal{R}^{61}$ where $P = (\alpha, \beta, \gamma)$. $\alpha \in \mathcal{R}^{10}$, $\beta \in \mathcal{R}^{45}$ and $\gamma \in \mathcal{R}^{6}$ represent the shape, articulation and global translation \& orientation respectively. Although loss minimization can be applied to directly optimize $P$ given the target 2D and 3D keypoints (obtained from manual annotation in multi-view setups) as in \cite{Zimmermann19}, 3D keypoints are not initially available in monocular RGB data. We find that loss computed using the target 2D keypoints alone oftentimes cannot reach a global minimum ${P}_{gt}$ from a default ${P}_{0} = \textbf{0}$ using gradient descent. However, with a proper ${P}_{0} = {P}_{gt} + \epsilon$ where $\epsilon \in \mathcal{R}^{61}$ represents an arbitrary error that is insufficient to deviate the gradient descent to an incorrect local minimum, we can successfully fit the hand model using 2D keypoints from a single RGB image.\\
\indent To allow effective manual modification of $P$, we first apply physical constraints to the pose space defined by $\beta \in \mathcal{R}^{45}$, which consists of 3 rotational values $(\theta, \phi, \psi)$ for each of the 15 finger joints and there are 3 joints $\textit{J}_{ij}$ defined for each finger $i$ where $i = [1, 5]$ and $j = [1, 3]$. Note that each joint of the original MANO model has 3 Degrees of Freedom (DoF) with unlimited range. This is obviously not realistic as all finger joints ${J}_{ij}$ primarily rotate in the direction of $\psi$ with ${J}_{i1}$ being able to rotate in the direction of $\theta$ as well (Figure \ref{fig:mano_samples}). Additionally, there is physical limitations for the range of rotation of each joint. To enforce physical plausibility, we define constant vectors ${\beta}_{min}$ and ${\beta}_{max}$ that clip $\beta$ within a reasonable range as follows, 
\begin{equation}\label{eq:beta_clipped}
{\beta}_{c} = min(max(\beta,\ {\beta}_{min}),\ {\beta}_{max}).
\end{equation}
To encourage more natural poses, we define ${\beta}_{mean} = ({\beta}_{min} + {\beta}_{max})/2$ and the following regularization loss,
\begin{equation}\label{eq:loss_manual_reg}
\mathcal{L}_{reg} = ({\beta}_{c} - {\beta}_{mean})^2\ \omega
\end{equation}
where $\omega \in \mathcal{R}^{45}$ applies element-wise scaling to the squared difference between the current pose and the mean pose. In general, $\mathcal{L}_{reg}$ punishes gradients for valid but less common rotations in the direction of $\theta$ for ${J}_{i0}$.\\
We formulate the ManoFit algorithm for manual annotation as the minimization problem below,
\begin{equation}\label{eq:loss_manual_2d}
\mathcal{L}_{2d} = \sum\limits_{k\in\mathcal{A}} (\bm{q}_{k} - \Pi(\bm{\tilde{p}}_{k}))^{2}
\end{equation}
\begin{equation}\label{eq:loss_fit_anno}
\mathcal{L}_{fit} = \mathcal{L}_{2d} + \mathcal{L}_{reg}
\end{equation}
where $\mathcal{L}_{2d}$ represents the Sum Squared Error (SSE) computed using the user-provided 2D keypoints $\bm{q} \in \mathcal{R}^{2}$ and the 2D projection $\Pi$ from 3D MANO keypoints $\bm{\tilde{p}} \in \mathcal{R}^{3}$ over the set of annotated joint indices $\mathcal{A}$. The combined loss $\mathcal{L}_{fit}$ aims to minimize the 2D keypoint error with valid and natural poses. Backpropagation using the Adam optimizer is applied to update $\beta$ and $\gamma$ with a stopping criteria that terminates when the loss ceases to improve for 10 iterations. Note that we use the default shape parameter $\alpha = \textbf{0}$. Although subject-specific $\alpha$ can improve fitting accuracy, shape-fitting requires multi-view data and our goal is to introduce a universal tool for hand pose annotation using monocular RGB. Qualitative and quantitative results in Section \ref{sec:fit_results} show that our fitting algorithm achieves excellent accuracy without subject-specific shape finetuning.\\
\indent For the annotation of each instance, the user is instructed to 1) modify $\gamma$ to the approximate values based on visualization of the MANO hand rendering, 2) annotate the wrist joint as well as the 5 finger tip joints, and 3) initiate parametric optimization. Until accurate fitting is achieved, the annotator can repeat the aforementioned steps with the freedom to modify $(\gamma, \beta)$ and annotate additional 2D joint locations. We demonstrate in the supplementary video that our annotation tool is well-designed for efficient control and fitting of the MANO hand model. Additional details are provided in the supplementary document.
\subsection{Annotated Data}
\begin{figure}[t]
  \small
  \centering
  \begin{subfigure}[t]{0.95\linewidth}
    \includegraphics[width=\linewidth]{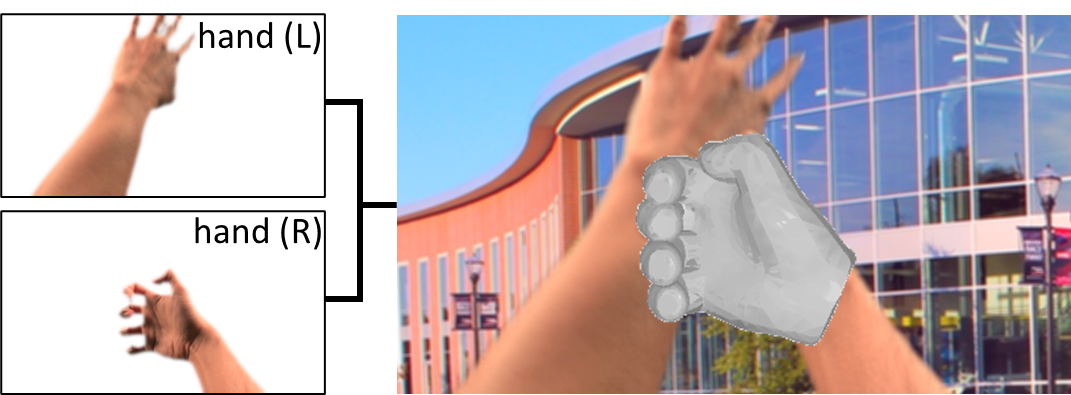}
  \end{subfigure}
  \caption{Illustration of two-hand image composition with visualized pose annotation of the primary right hand. }
  \label{fig:train_composition}
  \vspace{-4mm}
\end{figure}

\noindent\textbf{Training data.} Although the training set of Ego2Hands consists of$\sim180k$ frames, since some frames do not contain valid poses and some others have similar poses, we selected $7,033$ frames with diverse articulation from the training set of Ego2hands for manual annotation. For the training of 2D hand pose estimation model, we first follow~\cite{Lin20} and composite images at training-time. As illustrated in Figure \ref{fig:train_composition}, for each composited image, we randomly select the primary right hand from our annotated Ego2HandsPose training set, which contains the 3D hand pose annotation. For the secondary left hand, we randomly select a horizontally flipped right hand from the complete training set of Ego2Hands, which does not need the pose annotation as its purpose is to merely create a two-hand appearance. The background image is randomly selected from the proposed background set~\cite{Lin20}. For data augmentation, we apply random horizontal/vertical translation, color and smoothness augmentation. Quantitative evaluation in Section \ref{sec:quantitative_2d} shows that our composited data with pose annotation achieves state-of-the-art results on 2D hand pose estimation for our task.
\\
\noindent\textbf{Evaluation data.} As there is no evaluation benchmark for egocentric two-hand 3D global pose estimation in the wild, we manually annotate the complete test set of Ego2Hands, which consists of 8 diverse sequences with a total of $2,000$ two-hand images. In Section \ref{sec:quantitative_results}, we provide quantitative evaluation for two-hand 2D, 3D canonical and global hand pose estimation on this test set.
\section{2D Hand Pose Dataset Conversion}
\indent Using our annotation tool, we established in Section \ref{sec:manofit1} that manual guidance is needed to generate a well-initialized ${P}_{0}$ for parametric fitting on 2D keypoints from a single camera. In addition, 2D keypoints also need to be annotated but it is oftentimes unnecessary to annotate the complete set of 21 joints. In existing 2D hand pose datasets, since instances contain full annotation for the 2D keypoints, we can theoretically train a network to estimate the corresponding 3D canonical hand poses~\cite{Zimmermann17} and apply parametric fitting to generate the 3D hand pose annotations for 2D datasets. For the training of this network, as existing fixed-sized 3D hand pose datasets contain pose space with limited size that do not necessarily cover the true data distribution, we create a synthetic dataset MANO3DHands that provides the largest and most diverse pose space sufficient for accurate parametric fitting of any generic 2D hand pose dataset.
\subsection{MANO3DHands}
Since 3D canonical pose estimation models use heatmaps as input, there is no visual domain gap between heatmaps generated from synthetic and real-world data. However, synthetic hands with unrealistic articulation can lead to a domain gap in the pose space and a naively randomly sampled ${\beta}_{i}$ within the constrained space can still have an unrealistic combination of rotations.\\
\indent To sample realistic $\beta$ and $\gamma$, we first obtain two real-world data distributions of 3D hand poses from 5 participants using the LeapMotion device~\cite{Leapmotion}, which can automatically generate 3D keypoints from egocentric stereo infrared data. For the first distribution, we focus on $\gamma \in \mathcal{R}^{6}$ in the egocentric view and collect $31,796$ poses that cover a wide range of wrist rotations ${\gamma}_{1} \in \mathcal{R}^{3}$ and global locations ${\gamma}_{2} \in \mathcal{R}^{3}$. For the second distribution, we focus on $\beta$ and collect $42,496$ poses with diverse joint rotations. For both distributions with 3D keypoints, we apply parametric fitting to obtain the matching distributions $\mathcal{B}$ and $\mathcal{G}$ with MANO parameters for $\beta$ and $\gamma$ respectively.\\
\indent We find that egocentric and third-person viewpoints have different data distributions for the global orientation of the hand. In the egocentric viewpoint, we generate MANO hand poses by randomly sampling $(\beta, \gamma)$ from $\mathcal{B}$ and $\mathcal{G}$. In the third-person viewpoint, we sample $\gamma$ from all possible global orientations in $[-\pi, \pi]$. To enable evaluation, we generate two test sets with $50,000$ poses for each viewpoint. Section \ref{sec:quantitative_3d_can} shows that training on MANO3DHands achieves the best generalization score compared to existing 3D hand pose datasets in cross-dataset evaluation. 
\subsection{Unsupervised ManoFit}\label{sec:unsupervised_manofit}
With the complete set of annotated 2D keypoints $\bm{{q}_{gt}}$ as well as the estimated 3D canonical hand pose $\bm{{p}^{*}}$, we apply the multi-stage parametric fitting algorithm that progressively solves the following problems.\\
\noindent\textbf{1. Global orientation fitting.} We discovered in our experiments that the global orientation ${\gamma}_{1}$ has the greatest impact on the overall hand pose and should be properly optimized first using the following loss,
\begin{equation}\label{eq:loss_auto_gamma1}
\mathcal{L}_{{\gamma}_{1}} = {\lambda}_{{\gamma}_{1}}\ \frac{1}{N}\ \norm{L\cdot\bm{{p}^{*}} - \bm{\tilde{p}}}_{2}^{2}.
\end{equation}
For the number of joints $N = 21$, we minimize the Mean Squared Error (MSE) between the 3D keypoints $\bm{\tilde{p}}$ from the MANO model and the estimated 3D canonical keypoints $\bm{{p}^{*}}$ scaled by the reference bone length $L$. We use ${\lambda}_{{\gamma}_{1}} = \num{1e5}$ as the scaling constant.\\
\noindent\textbf{2. Pose articulation fitting.} After global orientation alignment, we optimize ${\gamma}_{1}$ and $\beta$ with the loss below,
\begin{equation}\label{eq:loss_auto_reg}
\mathcal{L}_{{\gamma}_{1}+\beta} = \mathcal{L}_{{\gamma}_{1}} + \mathcal{L}_{reg}
\end{equation}
where we use a combined loss of Equation \ref{eq:loss_auto_gamma1} and \ref{eq:loss_manual_reg} with ${\beta}_{mean}$ being the mean pose computed using the collected real-world pose distribution $\mathcal{B}$.\\
\noindent\textbf{3. Global translation fitting.} With $\beta$ and ${\gamma}_{1}$ properly aligned, we optimize ${\gamma}_{2}$ using $\mathcal{L}_{{\gamma}_{2}} = \mathcal{L}_{2d}$ defined in Equation \ref{eq:loss_manual_2d} with $\mathcal{A}$ being the complete set of joint indices.\\
\begin{figure*}[t]
  \small
  \centering
  \begin{subfigure}[t]{0.22\linewidth}
    \includegraphics[width=\linewidth]{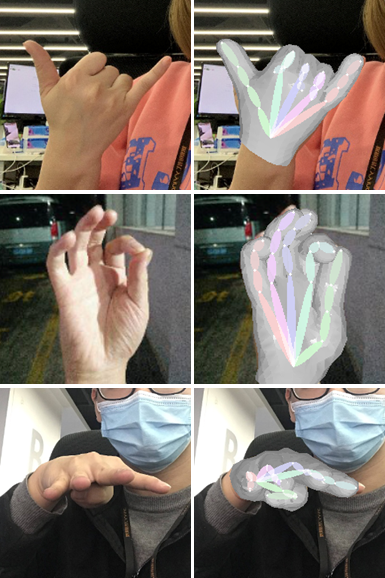}
    \caption{HIU-DMTL}
  \end{subfigure}
  \begin{subfigure}[t]{0.22\linewidth}
    \includegraphics[width=\linewidth]{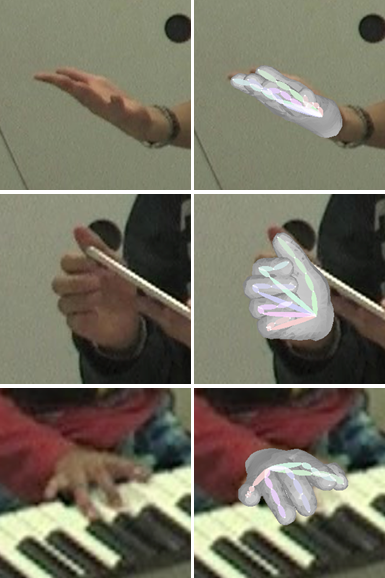}
    \caption{PanHand2D}
  \end{subfigure}
  \begin{subfigure}[t]{0.22\linewidth}
    \includegraphics[width=\linewidth]{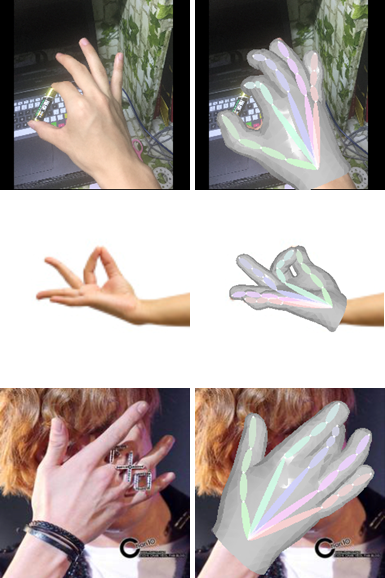}
    \caption{OneHand10k}
  \end{subfigure}
  \begin{subfigure}[t]{0.22\linewidth}
    \includegraphics[width=\linewidth]{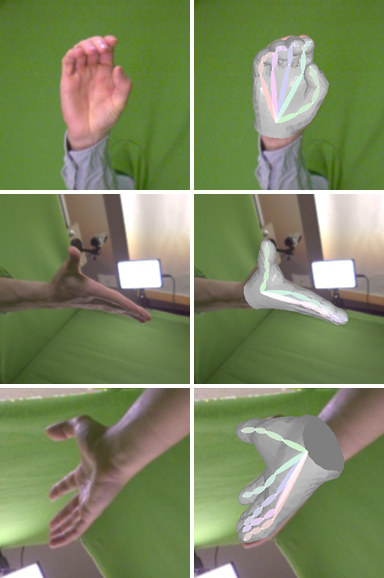}
    \caption{FreiHAND}
  \end{subfigure}
  \vspace{-2mm}
  \caption{Qualitative results of our generated 3D poses on 4 datasets using only 2D keypoints from a single image.}
  \label{fig:generated_datasets}
  \vspace{-4mm}
\end{figure*}
\begin{figure*}[t]
  \small
  \centering
  \begin{subfigure}[t]{1.00\linewidth}
    \includegraphics[width=\linewidth]{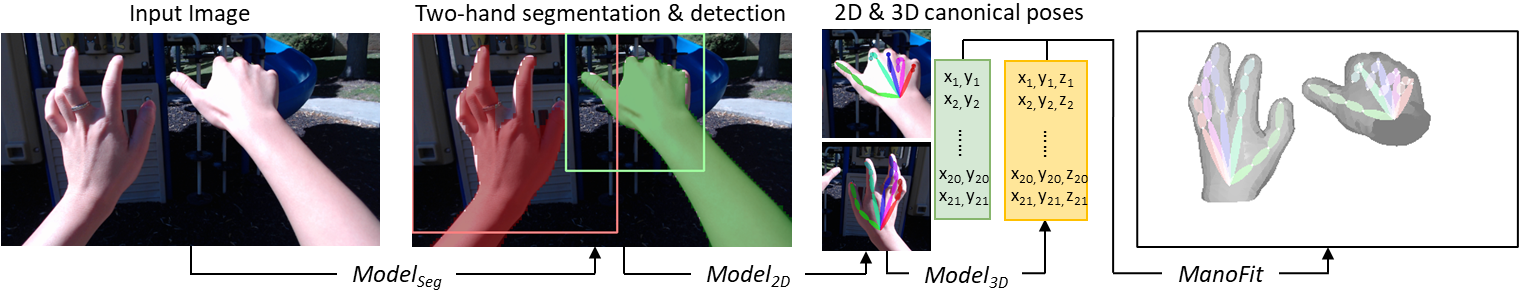}
  \end{subfigure}
  \vspace{-4mm}
  \caption{Overview of our two-hand 3D global pose estimation pipeline. Given an input image in the wild, we first extract hand bounding boxes and segmentation. 2D heatmaps are estimated using the cropped input and are used for 3D canonical pose estimation. MANO hand models in the global coordinate system are optimized using the 2D and 3D canonical poses.}
  \label{fig:twohandtrack_pipeline}
  \vspace{-4mm}
\end{figure*}
\noindent\textbf{4. Full pose fitting.} After the previous steps, we should have a well-initialized ${P}_{0} = {P}_{gt} + \epsilon$ for the final fitting. Therefore, we optimize $\beta$ and $\gamma$ using $\mathcal{L}_{fit} = \mathcal{L}_{2d} + \mathcal{L}_{reg}$. \\
\indent We perform optimization for 100 and 300 iterations for stages 1-3 and 4 respectively. The learning rate is set to $1.0$, $0.01$, $1.0$ and $0.01$ for stage 1-4. This process can achieve accurate matching between the ground truth 2D keypoints $\bm{{q}_{gt}}$ and the projected 2D keypoints $\Pi(\bm{\tilde{p}})$ from the 3D keypoints of the MANO hand model.
\subsection{Fitting Results \& Analysis}\label{sec:fit_results}
\indent We select the following 2D hand pose datasets for automatic conversion: HIU-DMTL~\cite{Zhang21} (41,539 instances), PanHand2D~\cite{Simon17} (14,817 instances) and OneHand10k~\cite{Wang18} (2,040 instances). Instances without full 2D annotation in OneHand10k are discarded. In addition, since hand side information is necessary for parametric fitting, we manually annotate hand side labels for HIU-DMTL and OneHand10k, which contain both left and right hand instances.\\
\indent Qualitative examples in Figure \ref{fig:generated_datasets} show that our algorithm can accurately fit a wide range of poses from different datasets. To quantitatively evaluate our fitting algorithm, we generate 3D hand poses for the training set of FreiHAND, which provides $32,560$ annotated instances with diverse 3D hand poses. We apply scaling using the reference bone length and Cartesian alignment using the ground truth absolute 3D location of the root joint. Our automatic fitting achieves an End Point Error (EPE) of $1.17$cm. \\
\indent Similar to our parametric fitting algorithm,~\cite{Kulon20} proposed to automatically fit the MANO hand model using 2D keypoints estimated using OpenPose~\cite{Cao17} without using the estimated 3D canonical keypoints. We point out that this approach only selects samples that pass the designed heuristic verification and is limited to fit poses with ${P}_{0}$ less susceptible to local minima during gradient descent. In comparison, after manual multi-view validation for all generated instances using our algorithm, we report high acceptance rates of $84.6\%$, $88.1\%$ and $79.1\%$ for HIU-DMTL, PanHand2D and OneHand10k respectively. Note that a small rejection rate is expected due to the depth ambiguity in 2D keypoints.
\section{Two-hand 3D Global Pose Estimation}
\indent For comprehensive analysis on our proposed dataset, we follow~\cite{Lin19} and use a multi-stage pipeline for the task of two-hand 3D global pose estimation (Figure \ref{fig:twohandtrack_pipeline}). \\
\noindent\textbf{Two-hand segmentation and detection.} For the first stage, We use the scene-adapted ICNet from~\cite{Lin20} to estimate the segmentation as well as the activation energy for both hands. Segmentation is used to address inter-hand occlusion and hand energy that excludes the arm is used for hand detection even when occluded. We apply a binary threshold using $\tau = 0.5$ to the estimated energy and perform a close operation with kernel size of 3 for noise removal.\\
\noindent\textbf{2D hand pose estimation.} Using the energy mask, we obtain the cropped images for 2D hand pose estimation. In addition to the three-channel RGB input, we concatenate the cropped binary segmentation mask for the other hand to encode occlusion information. We train HRNet-W32~\cite{Sun19} on the training set of Ego2HandsPose with a batch size of $16$ for $40,000$ iterations. The Adam optimizer is used with an initial learning rate of $0.0001$ (decreasing with a rate of 0.5 per $10,000$ iterations). Using input images resized to $224\times224$, the output heatmaps have a resolution of $56\times56$. To simplify the pose space, we horizontally flip the left hand so the network trains on the right hand only.\\
\noindent\textbf{3D canonical hand pose estimation.} We use a compact ResNet10 with 3 fully connected layers to regress the root-relative 3D joint locations. We generate training instances online using data distributions $\mathcal{B}$ and $\mathcal{G}$ from the proposed MANO3DHands and train for 400k iterations with a batch size of 1 and a learning rate that decreases with a rate of 0.5 per 100k iterations.\\
\noindent\textbf{Hand tracking via Manofit.} For each hand's reappearance, we first use the projection algorithm proposed by ~\cite{Lin19} to set the global translation parameter ${\gamma}_{2}$. The Manofit algorithm introduced in Section \ref{sec:unsupervised_manofit} is then applied for hand tracking. For continuous tracking, as subsequent frames contain poses with gradual changes, we do not reset the MANO parameters $P$ or the internal state of the Adam optimizer prior to optimization, which conveniently leads to temporally smooth pose estimation. For each optimization stage, a properly selected threshold value is used on the corresponding loss as the stopping criteria.
\begin{figure}[t]
  \small
  \centering
  \begin{subfigure}[t]{0.95\linewidth}
    \includegraphics[width=\linewidth]{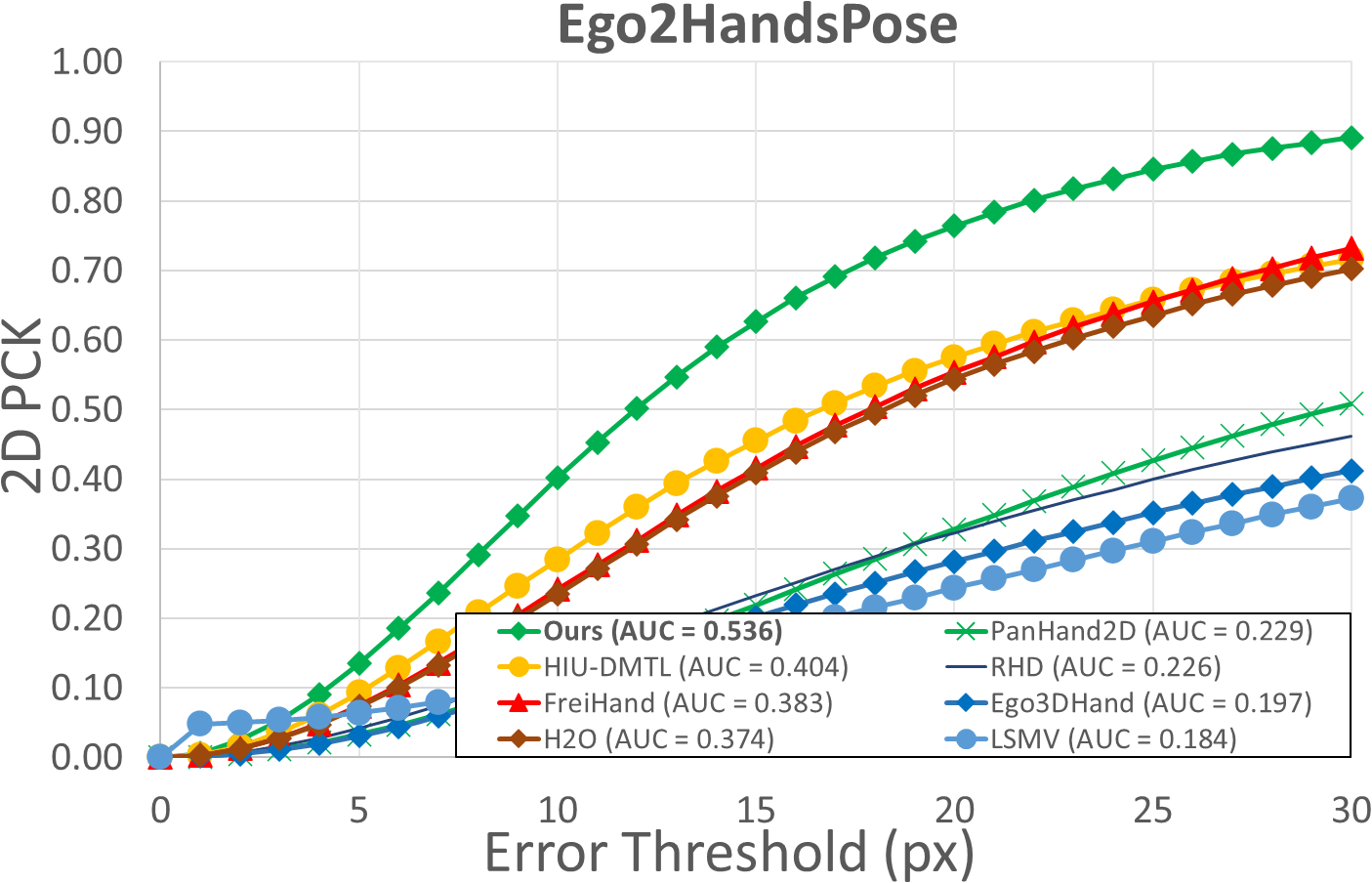}
  \end{subfigure}
  \caption{Quantitative comparison of 2D hand pose estimation on the proposed dataset. Image resolution of $800\times448$ is used.}
  \label{fig:eval_2d_ce}
  \vspace{-4mm}
\end{figure}
\section{Quantitative Benchmarking}\label{sec:quantitative_results}
\indent In this section, we demonstrate that Ego2HandsPose enables state-of-the-art results on egocentric two-hand 3D global pose estimation in the wild. First, we evaluate on the test set of Ego2HandsPose and perform isolated studies on each stage of our proposed pipeline. Second, we evaluate using the complete pipeline on its ability to track both hands in the global coordinate system.
\subsection{2D Hand Pose Estimation}\label{sec:quantitative_2d}
\indent We use the ground truth two-hand energy to obtain the cropped hand images as input and evaluate using HRNet-W32 trained on 8 large-scale datasets with data augmentation. Figure \ref{fig:eval_2d_ce} shows that training on our dataset significantly outperforms others and enables the top $\text{AUC}_{2D} = 0.536$. We theorize the performance gap is caused by the difference in pose space (third-person for HIU-DMTL, FreiHAND) and the lack of diversity (H2O). Note that datasets with synthetic data or laboratory backgrounds have the lowest accuracy. Although it is expected for a network to perform well on the dataset it was trained on, we argue that since the test set of Ego2Hands contain subjects and scenes not present in its training set, the score obtained by Ego2HandsPose demonstrates its ability to enable accurate estimation on unseen data. Despite the fact that cross-dataset evaluation is commonly performed to show the generalization ability of a dataset, we do not perform this experiment in this stage as it is not our goal to generalize to third-person or synthetic visual data, but to achieve the best accuracy on the task of egocentric two-hand tracking.
\subsection{3D Canonical Hand Pose Estimation}\label{sec:quantitative_3d_can}
\indent In this stage, we use heatmaps as input and evaluate cross-dataset performance using 6 large-scale datasets with diverse 3D hand pose annotations, including our proposed MANO3DHands with egocentric and third-person data distributions. The complete datasets are used in training and evaluation for more comprehensive analysis. As a result, the non-diagonal scores should be emphasized to analyze the generalization ability, which correlates with the pose space quality each dataset provides. Table \ref{tab:table_cross_eval} shows that our proposed MANO3DHands datasets achieve the highest generalization scores with average rankings of $2.3$. This finding indicates that the generated pose space using our collected distributions ($\mathcal{B}$ and $\mathcal{G}$) better represent the true pose space compared to other datasets. In addition, it is important to recognize the difference between egocentric and third-person pose space. We show that $\text{MANO3DHands}_{3rd}$ and $\text{MANO3DHands}_{ego}$ achieve the best generalization scores for third-person and egocentric datasets respectively. Therefore, we claim that it is best to train on hand pose data with the matching viewpoints for different applications. In this work, we use $\text{MANO3DHands}_{3rd}$ for the annotation tool and $\text{MANO3DHands}_{ego}$ for egocentric two-hand tracking.\\
\indent Unlike traditional datasets with fixed sizes, the training set of MANO3DHands dynamically generates instances using the collected distribution and does not have a static size. For this reason, the represented pose space can be significantly higher in quantity and diversity. For example, $\text{MANO3DHands}_{ego}$ can generate $\num{1.35e9}$ unique poses with the collected $\mathcal{B}$ and $\mathcal{G}$. Consequently, the top evaluation scores achieved by our datasets on our generated test sets (50k instances) also reflect strong generalization ability.
\begin{table}[t]
  \small
  \centering
  \begin{tabular}{P{0.9cm}||c|c|c|c|c|c}
  \toprule
Dataset 	 &
Ours&
Ours*&
H2O*&
Frei&
LSMV&
Ego3D*
\\\hline\hline
Ours&\textbf{0.735}&{\color{blue}0.732}& 0.743 & {\color{blue}0.731} & {\color{blue}0.613} & {\color{ForestGreen}0.776}\\\hline
Ours*&{\color{ForestGreen}0.611}&\textbf{0.790}& \textbf{0.827} & {\color{ForestGreen}0.643} & 0.522 & {\color{blue}0.790}\\\hline
H2O*&0.473&0.655& {\color{blue}0.821} & 0.494 & 0.412 & 0.671\\\hline
Frei&{\color{blue}0.686}&{\color{ForestGreen}0.705}& {\color{ForestGreen}0.744} & \textbf{0.738} & {\color{ForestGreen}0.607} & 0.743\\\hline
LSMV&0.528&0.482& 0.490 & 0.541 & \textbf{0.646} & 0.493\\\hline
Ego3D*&0.523&0.679& 0.688 & 0.551 & 0.404 & \textbf{0.834}\\\hline
Rank& 2.3 & 2.3 & 4.8 & 2.7 & 4.7 & 4.2
\end{tabular}
\caption{Cross-dataset evaluation on 3D canonical pose estimation. We report AUC computed using PCK of root-relative 3D keypoints in an interval from $0.0$ to $1.0$. Egocentric datasets are labeled with *. The top 3 scores on each evaluation dataset (shown in columns) are marked as {\color{black}\textbf{first}}, {\color{blue}second} and {\color{ForestGreen}third}. }
\label{tab:table_cross_eval}
\vspace{-5mm}
\end{table}
\begin{figure*}[t]
  \small
  \centering
  \begin{subfigure}[t]{0.33\linewidth}
    \includegraphics[width=\linewidth]{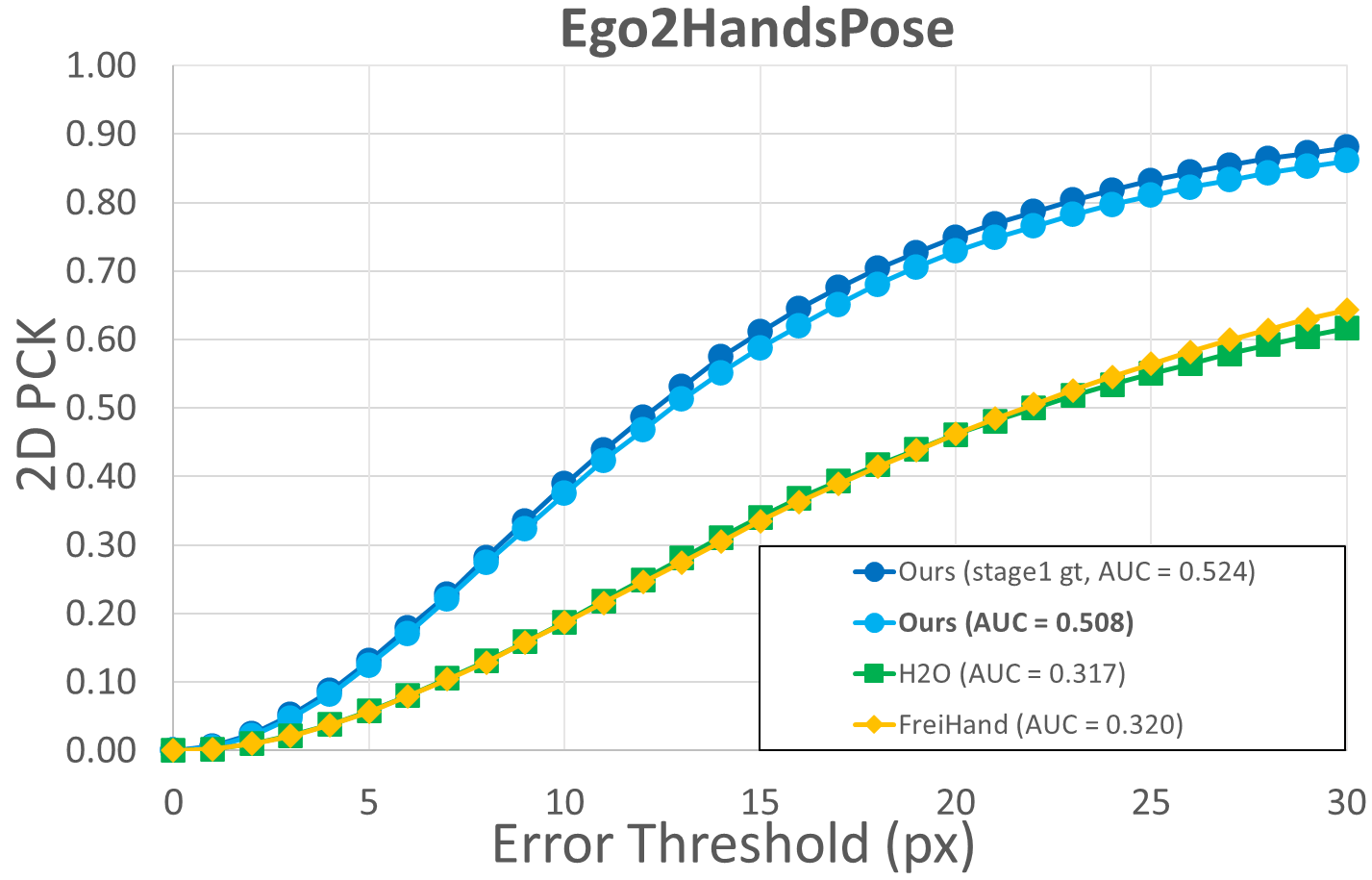}
    \caption{2D Pose Estimation}
    \label{fig:eval_3d_glog_2d}
  \end{subfigure}
  \begin{subfigure}[t]{0.33\linewidth}
    \includegraphics[width=\linewidth]{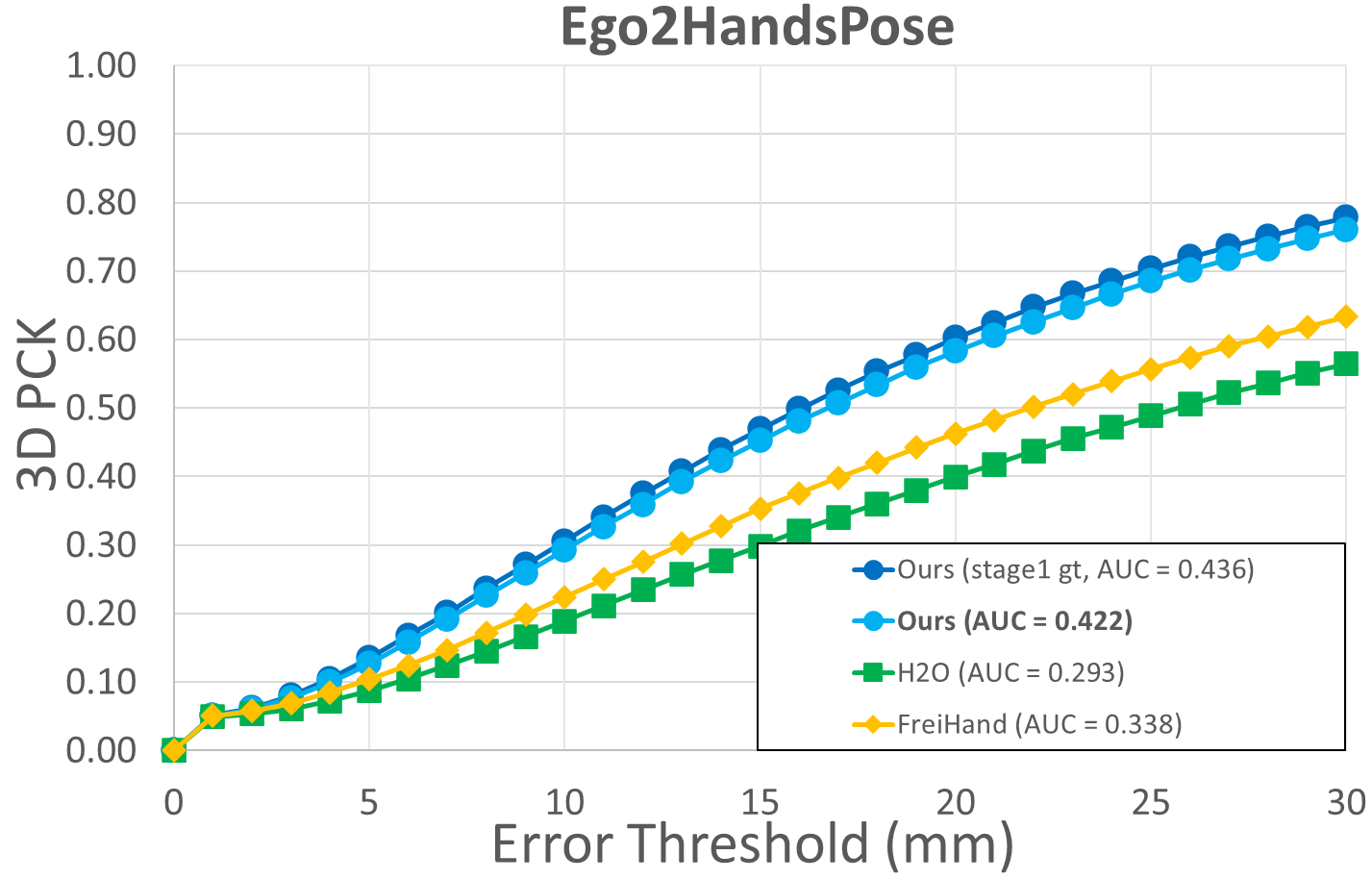}
    \caption{3D Canonical Pose Estimation}
    \label{fig:eval_3d_glog_3d_can}
  \end{subfigure}
  \begin{subfigure}[t]{0.33\linewidth}
    \includegraphics[width=\linewidth]{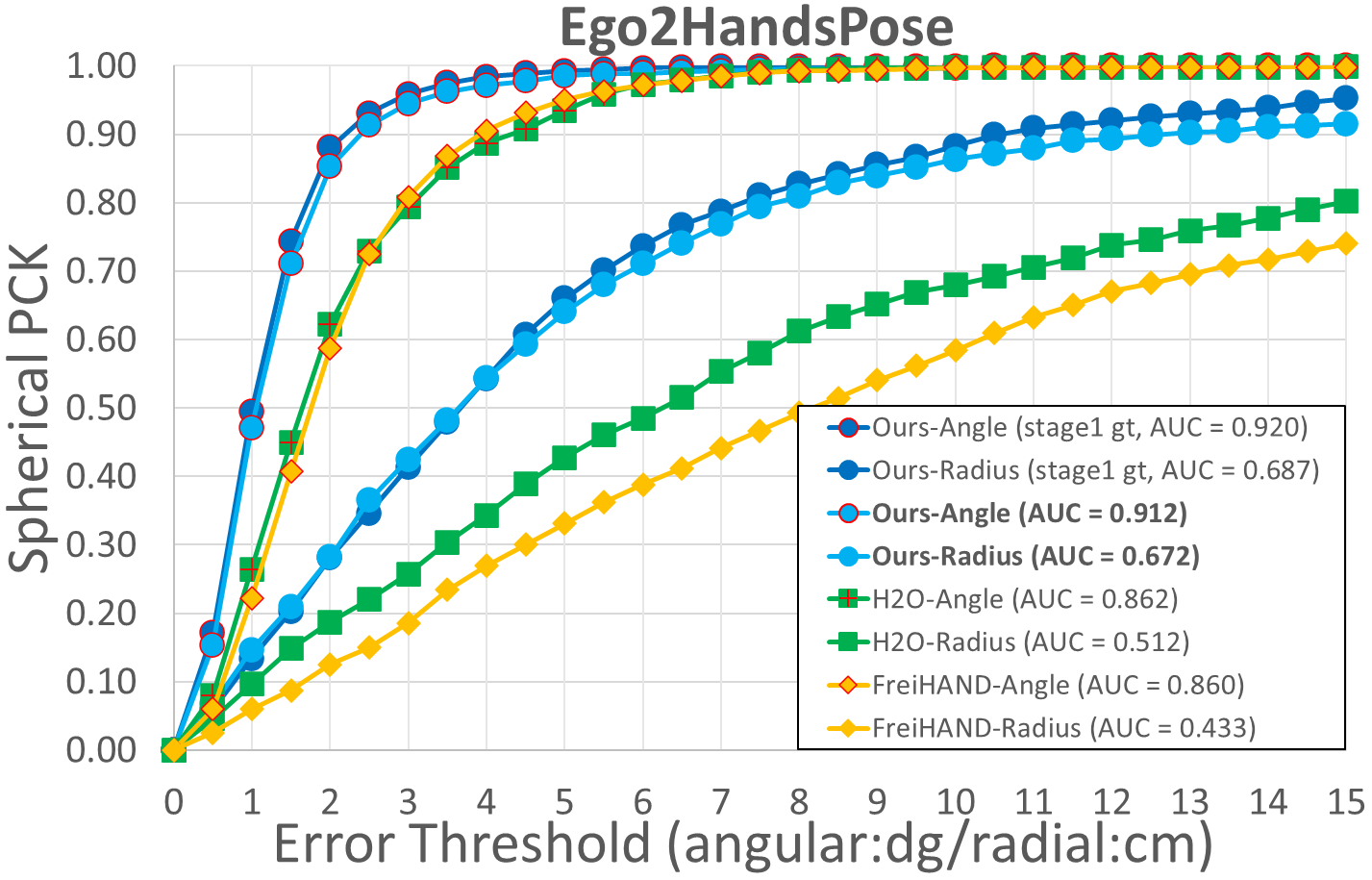}
    \caption{Spherical PCK for Global Estimation}
    \label{fig:eval_3d_glog_spherical}
  \end{subfigure}
  \vspace{-2mm}
  \caption{Quantitative results using the complete pipeline trained on different datasets. Results obtained using the ground truth segmentation and detection (stage1 ground truth) are provided to study the impact of $\text{Model}_{seg}$ on the overall accuracy.}
  \label{fig:eval_3d_glob}
  \vspace{-2mm}
\end{figure*}
\begin{figure*}[t]
  \small
  \centering
  \begin{subfigure}[t]{0.24\linewidth}
    \includegraphics[width=\linewidth]{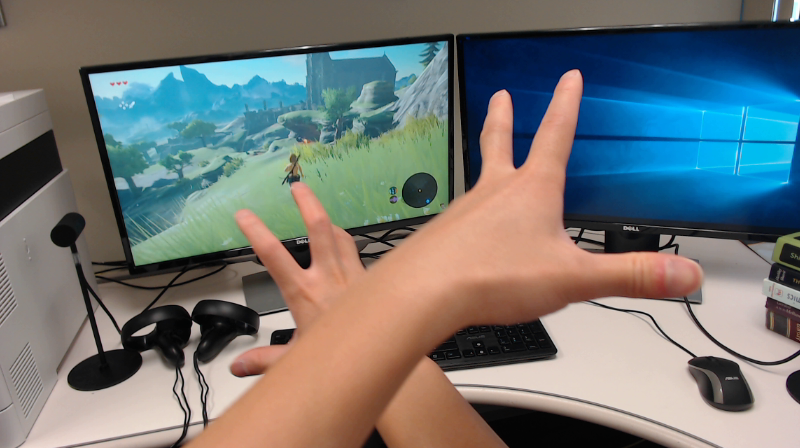}
  \end{subfigure}
  \begin{subfigure}[t]{0.24\linewidth}
    \includegraphics[width=\linewidth]{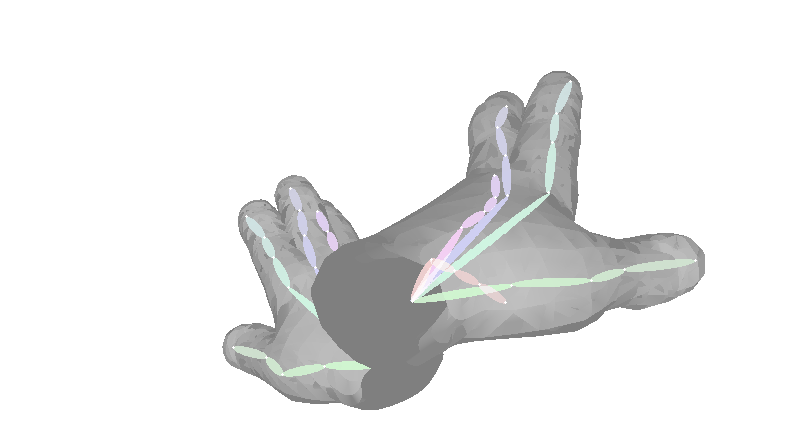}
  \end{subfigure}
  \begin{subfigure}[t]{0.24\linewidth}
    \includegraphics[width=\linewidth]{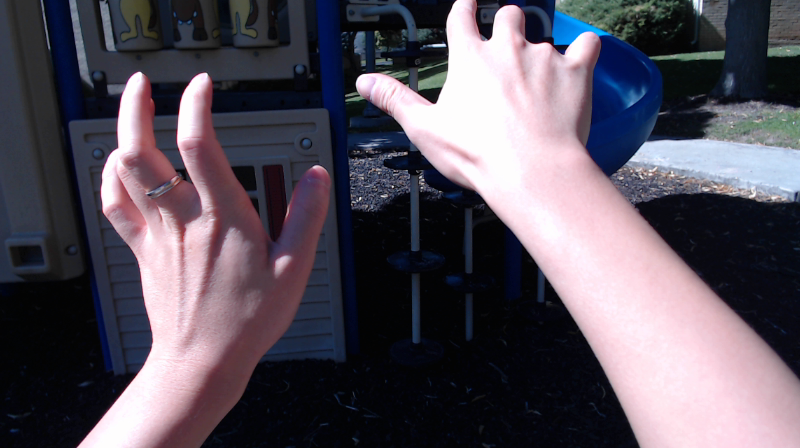}
  \end{subfigure}
  \begin{subfigure}[t]{0.24\linewidth}
    \includegraphics[width=\linewidth]{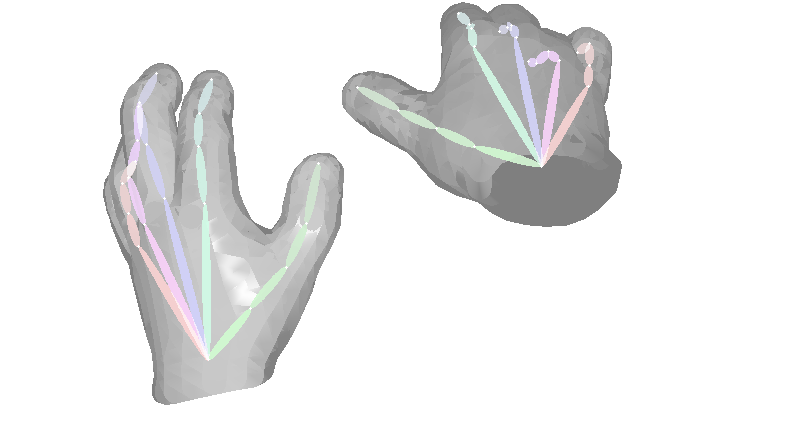}
  \end{subfigure}
  \begin{subfigure}[t]{0.24\linewidth}
    \includegraphics[width=\linewidth]{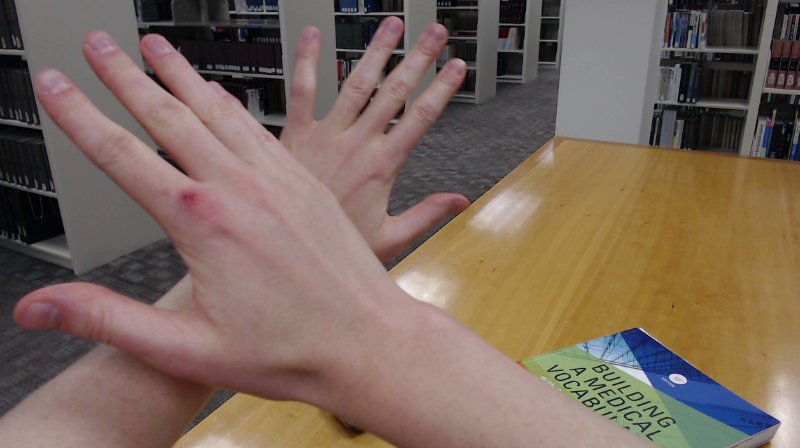}
  \end{subfigure}
  \begin{subfigure}[t]{0.24\linewidth}
    \includegraphics[width=\linewidth]{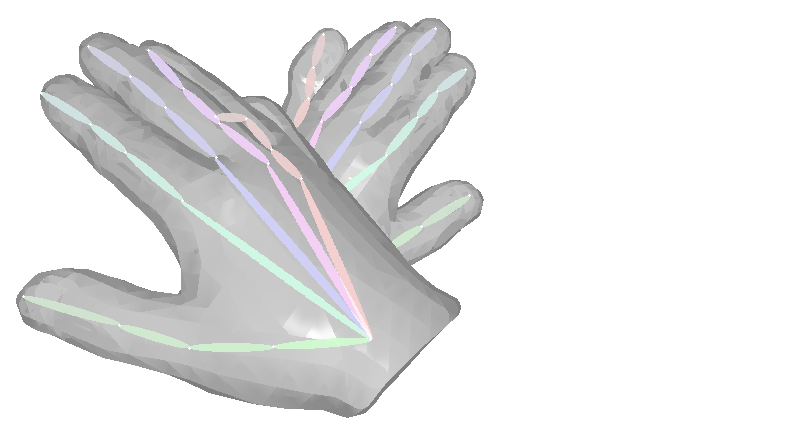}
  \end{subfigure}
  \begin{subfigure}[t]{0.24\linewidth}
    \includegraphics[width=\linewidth]{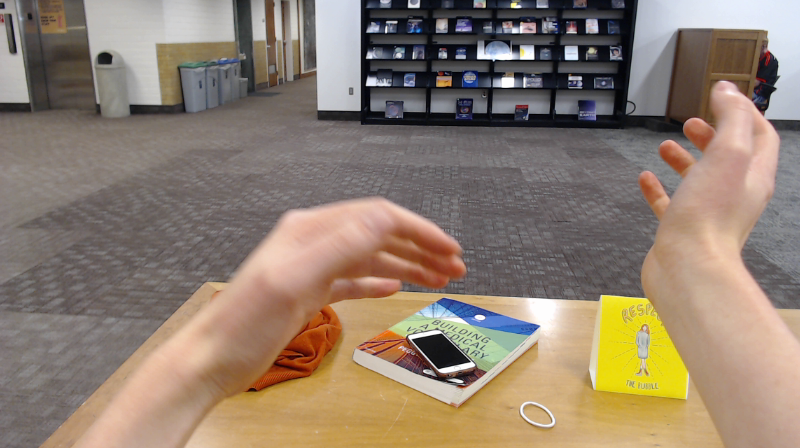}
  \end{subfigure}
  \begin{subfigure}[t]{0.24\linewidth}
    \includegraphics[width=\linewidth]{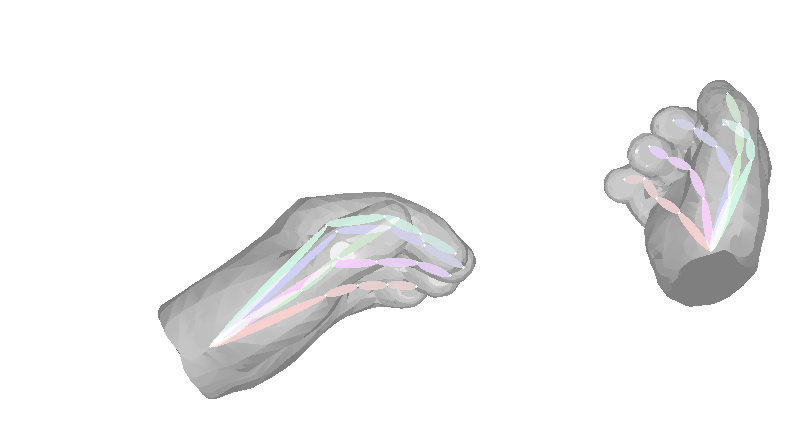}
  \end{subfigure}
  \begin{subfigure}[t]{0.24\linewidth}
    \includegraphics[width=\linewidth]{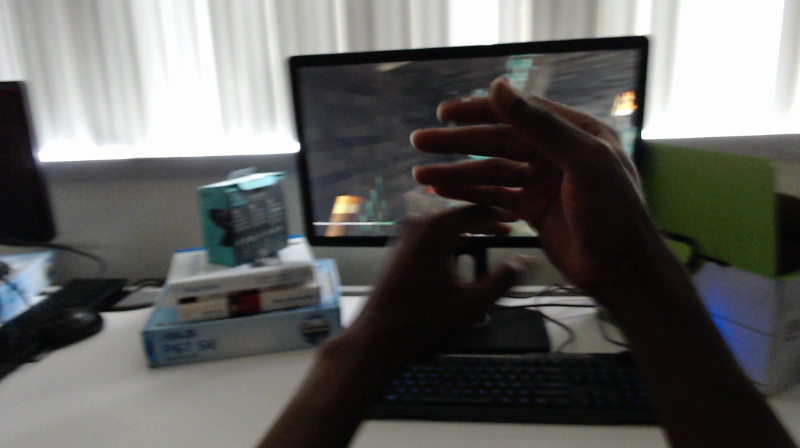}
  \end{subfigure}
  \begin{subfigure}[t]{0.24\linewidth}
    \includegraphics[width=\linewidth]{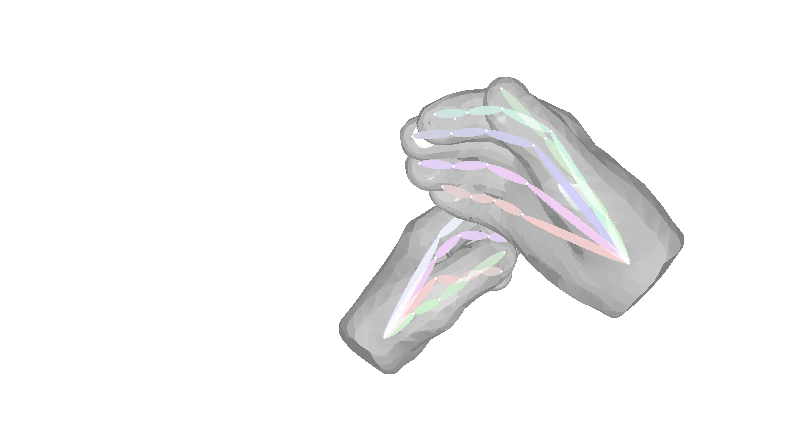}
  \end{subfigure}
  \begin{subfigure}[t]{0.24\linewidth}
    \includegraphics[width=\linewidth]{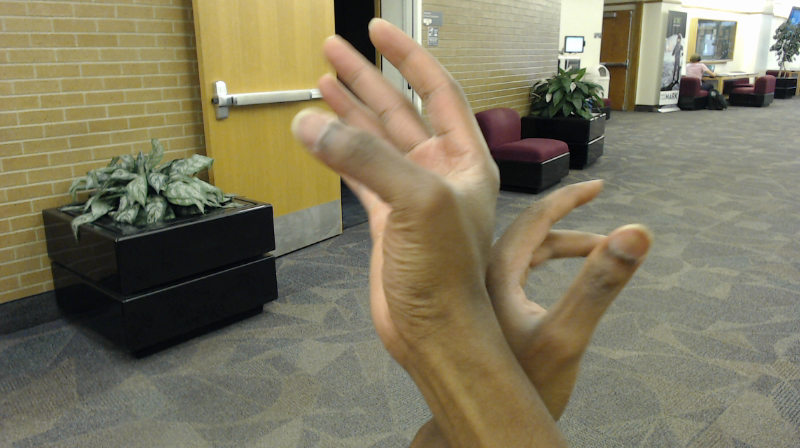}
  \end{subfigure}
  \begin{subfigure}[t]{0.24\linewidth}
    \includegraphics[width=\linewidth]{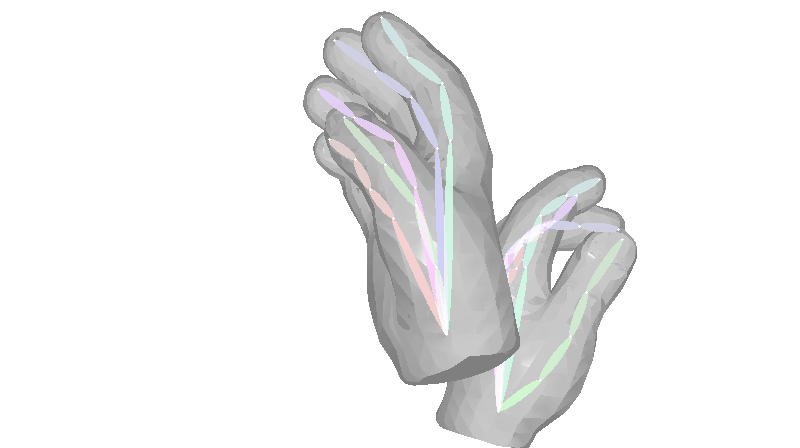}
  \end{subfigure}
  \begin{subfigure}[t]{0.24\linewidth}
    \includegraphics[width=\linewidth]{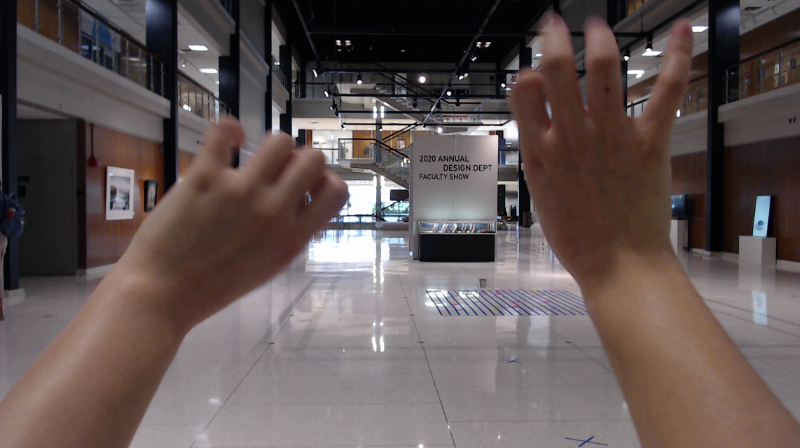}
  \end{subfigure}
  \begin{subfigure}[t]{0.24\linewidth}
    \includegraphics[width=\linewidth]{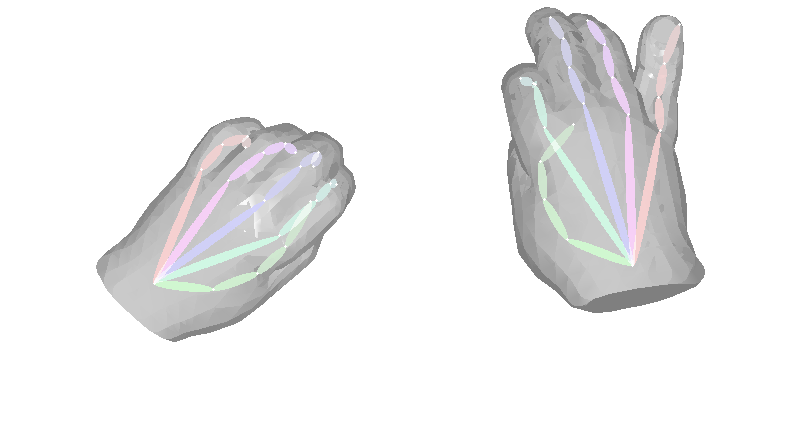}
  \end{subfigure}
  \begin{subfigure}[t]{0.24\linewidth}
    \includegraphics[width=\linewidth]{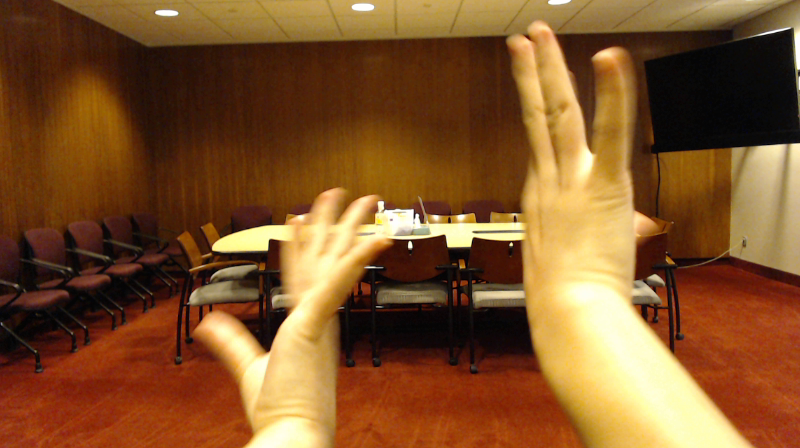}
  \end{subfigure}
  \begin{subfigure}[t]{0.24\linewidth}
    \includegraphics[width=\linewidth]{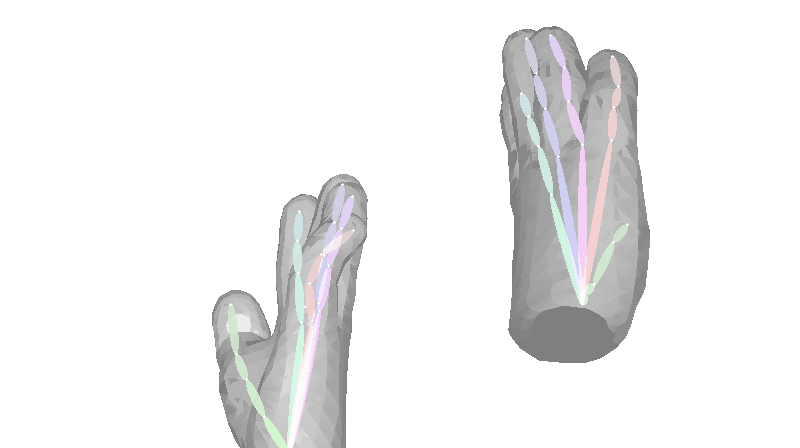}
  \end{subfigure}
  \caption{Qualitative results obtained using our multi-stage method trained on Ego2HandsPose. Odd columns show the input images from various sequences in the test set of Ego2HandsPose. Even columns show visualization of the ManoFit output.}
  \label{fig:qualitative_results}
  \vspace{-4mm}
\end{figure*}
\subsection{Two-hand Global Pose Estimation}\label{sec:quantitative_3d_glob}
\indent For the complete cascaded pipeline, we use the scene-adapted ICNet and evaluate subsequent stages using input obtained from the previous stages. For 3D global hand pose estimation that requires the global 3D hand location, we follow~\cite{Lin19} and compute the PCK on the root joint in the spherical coordinate system, which measures the directional and distance accuracy. To show that existing datasets are insufficient for our task, we select models trained on H2O and FreiHAND with good performance in Section \ref{sec:quantitative_2d} and \ref{sec:quantitative_3d_can} for a comparison in the complete pipeline.\\
\indent Figure \ref{fig:eval_3d_glog_2d} shows that we achieve an $\text{AUC}_{2d} = 0.508$ for 2D hand pose estimation. Accurate results in this stage are necessary for two-hand tracking since both the third and final stage heavily depend on the estimated 2D keypoints. For 3D canonical hand pose estimation, we achieve an $\text{AUC}_{3d} = 0.422$ in Figure \ref{fig:eval_3d_glog_3d_can}. Finally, after applying ManoFit using the estimated 2D/3D hand poses, Figure \ref{fig:eval_3d_glog_spherical} shows that we obtain top $\text{AUC}_{angle} = 0.912$ and $\text{AUC}_{radius} = 0.672$. To isolate the impact of $\text{Model}_{seg}$ in the first stage, we also provide results obtained using the ground truth segmentation and detection, which is not used in comparison with the other datasets. Figure \ref{fig:qualitative_results} shows qualitative examples of our two-hand 3D tracking on the test sequences of Ego2HandsPose. Additional results are provided in the supplementary material.\\
\indent We note that there is a trade-off between fitting accuracy and inference time and our complete pipeline does not currently run in real-time. We report an average inference time of $19.7$ms, $50.1$ms and $2.4$ms for our selected $\text{Model}_{seg}$, $\text{Model}_{2d}$ and $\text{Model}_{3d}$ respectively. We utilize high-performance models for the challenging task of two-hand tracking using a single camera in the wild. Future work includes further optimizations in efficiency.
\section{Conclusion}
In this work, we propose a set of parametric fitting algorithm that enables 3D hand pose annotation using a single image and automatic conversion from 2D to 3D hand poses. We propose the first dataset, Ego2HandsPose, that tackles two-hand 3D global pose estimation in the wild using a monocular RGB. Results obtained using our multi-stage pipeline shows that training on the proposed dataset significantly outperforms existing datasets. We hope our work can push color-based two-hand applications towards unconstrained environments for practical applications.

{\small
\bibliographystyle{ieee_fullname}
\bibliography{egbib}
}

\clearpage
\onecolumn
\setcounter{figure}{0}
\setcounter{table}{0}
\begin{center}
{\Large\textbf{Supplementary Document:\\
Ego2HandsPose: A Dataset for Egocentric Two-hand 3D Global Pose Estimation}}
\end{center}
\section{ManoFit Annotation Tool}
In this section, we provide additional details for the introduced hand pose annotation tool, which consists of three major panels.
\begin{figure*}[h]
  \small
  \centering
  \begin{subfigure}[t]{0.95\linewidth}
    \includegraphics[width=\linewidth]{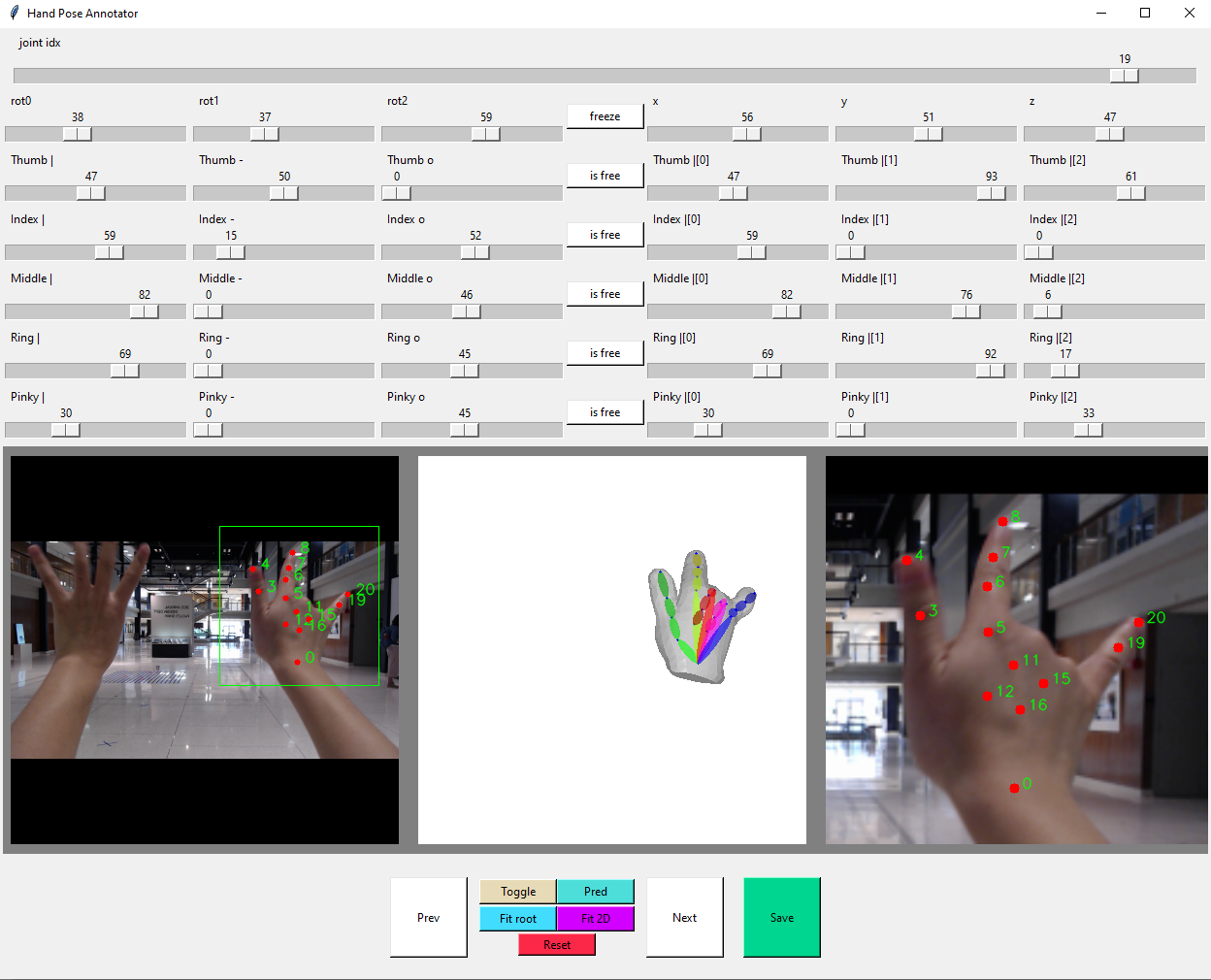}
  \end{subfigure}
  \caption{Hand pose annotation tool for ManoFit using a single image.}
  \label{fig:annotation_tool}
  \vspace{-4mm}
\end{figure*}
\subsection{Display Panel}
\indent As shown in Figure \ref{fig:annotation_tool}, we display three images for annotation. The center image displays the rendered MANO hand model with a resolution of $400\times400$. The left image consists of the input image resized and padded to match the same resolution for fitting. The user has the option to crop a hand region for a magnified view (the right image) or automatic 2D/3D hand pose estimation using our pretrained models. 2D joint locations can be annotated by first selecting the joint using the top slider and then clicking on the left or the right image.
\subsection{Control Panel}
We provide sliders to control global orientation ${\gamma}_{1}$ and translation ${\gamma}_{2}$. In addition, sliders for the control of each finger is provided. While the first joint of the selected finger has degrees of freedom to rotate in the directions of $\theta$, $\phi$ and $\psi$ with the predefined ranges, the second and third joint of each finger can only rotate in the direction of $\psi$. This panel allows the user to have full manual control of the MANO hand model.
\subsection{Action Panel}
With an arbitrary number of provided 2D joint locations, the user can apply our parametric fitting algorithm to automatically fit MANO parameters using the defined losses. To automatically obtain the 2D joint locations, we allow the user to estimate the 2D/3D canonical hand poses using our pretrained models. A toggle button is provided to assist the user with improved ability to visualize how well the rendered hand model matches the input image. Other basic functionalities such as navigation between input images, resetting the MANO pose and saving MANO parameters are also provided.
\begin{table*}[t!]
  \small
  \centering
  \begin{tabular}{c|c|c|c|c}
  \toprule
Sequence 	&
\begin{tabular}{c}$\text{AUC}_{2D}$\\\hline
    	\begin{tabular}{c|P{0.9cm}|c}Ours & H2O & Frei\end{tabular}
\end{tabular} & 
\begin{tabular}{c}$\text{AUC}_{3D}$\\\hline
    	\begin{tabular}{c|P{0.9cm}|c}Ours & H2O & Frei\end{tabular}
\end{tabular} & 
\begin{tabular}{c}$\text{AUC}_{angle}$\\\hline
    	\begin{tabular}{c|P{0.9cm}|c}Ours & H2O & Frei\end{tabular}
\end{tabular} &
\begin{tabular}{c}$\text{AUC}_{radius}$\\\hline
    	\begin{tabular}{c|P{0.9cm}|c}Ours & H2O & Frei\end{tabular}
\end{tabular}\\
\midrule
seq-1&\begin{tabular}{P{0.8cm}|P{0.9cm}|c}\textbf{0.574}&0.476&0.464\end{tabular}&\begin{tabular}{P{0.8cm}|P{0.9cm}|c}\textbf{0.477}&0.338&0.377\end{tabular}&\begin{tabular}{P{0.8cm}|P{0.9cm}|c}\textbf{0.938}&0.913&0.907\end{tabular}&\begin{tabular}{P{0.8cm}|P{0.9cm}|c}\textbf{0.786}&0.658&0.581\end{tabular}\\
seq-2&\begin{tabular}{P{0.8cm}|P{0.9cm}|c}\textbf{0.475}&0.095&0.136\end{tabular}&\begin{tabular}{P{0.8cm}|P{0.9cm}|c}\textbf{0.378}&0.149&0.203\end{tabular}&\begin{tabular}{P{0.8cm}|P{0.9cm}|c}\textbf{0.917}&0.770&0.762\end{tabular}&\begin{tabular}{P{0.8cm}|P{0.9cm}|c}\textbf{0.741}&0.192&0.078\end{tabular}\\
seq-3&\begin{tabular}{P{0.8cm}|P{0.9cm}|c}\textbf{0.564}&0.429&0.319\end{tabular}&\begin{tabular}{P{0.8cm}|P{0.9cm}|c}\textbf{0.518}&0.415&0.415\end{tabular}&\begin{tabular}{P{0.8cm}|P{0.9cm}|c}\textbf{0.943}&0.924&0.865\end{tabular}&\begin{tabular}{P{0.8cm}|P{0.9cm}|c}\textbf{0.759}&0.747&0.547\end{tabular}\\
seq-4&\begin{tabular}{P{0.8cm}|P{0.9cm}|c}\textbf{0.581}&0.480&0.432\end{tabular}&\begin{tabular}{P{0.8cm}|P{0.9cm}|c}\textbf{0.499}&0.438&0.474\end{tabular}&\begin{tabular}{P{0.8cm}|P{0.9cm}|c}\textbf{0.936}&0.925&0.907\end{tabular}&\begin{tabular}{P{0.8cm}|P{0.9cm}|c}\textbf{0.846}&0.765&0.731\end{tabular}\\
seq-5&\begin{tabular}{P{0.8cm}|P{0.9cm}|c}\textbf{0.516}&0.270&0.289\end{tabular}&\begin{tabular}{P{0.8cm}|P{0.9cm}|c}\textbf{0.417}&0.278&0.282\end{tabular}&\begin{tabular}{P{0.8cm}|P{0.9cm}|c}\textbf{0.911}&0.858&0.843\end{tabular}&\begin{tabular}{P{0.8cm}|P{0.9cm}|c}\textbf{0.660}&0.315&0.211\end{tabular}\\
seq-6&\begin{tabular}{P{0.8cm}|P{0.9cm}|c}\textbf{0.511}&0.374&0.278\end{tabular}&\begin{tabular}{P{0.8cm}|P{0.9cm}|c}\textbf{0.499}&0.425&0.411\end{tabular}&\begin{tabular}{P{0.8cm}|P{0.9cm}|c}\textbf{0.926}&0.899&0.856\end{tabular}&\begin{tabular}{P{0.8cm}|P{0.9cm}|c}\textbf{0.825}&0.612&0.472\end{tabular}\\
seq-7&\begin{tabular}{P{0.8cm}|P{0.9cm}|c}\textbf{0.574}&0.255&0.396\end{tabular}&\begin{tabular}{P{0.8cm}|P{0.9cm}|c}\textbf{0.362}&0.165&0.310\end{tabular}&\begin{tabular}{P{0.8cm}|P{0.9cm}|c}\textbf{0.923}&0.828&0.902\end{tabular}&\begin{tabular}{P{0.8cm}|P{0.9cm}|c}\textbf{0.760}&0.460&0.510\end{tabular}\\
seq-8&\begin{tabular}{P{0.8cm}|P{0.9cm}|c}\textbf{0.375}&0.155&0.246\end{tabular}&\begin{tabular}{P{0.8cm}|P{0.9cm}|c}\textbf{0.303}&0.137&0.232\end{tabular}&\begin{tabular}{P{0.8cm}|P{0.9cm}|c}\textbf{0.899}&0.782&0.840\end{tabular}&\begin{tabular}{P{0.8cm}|P{0.9cm}|c}\textbf{0.693}&0.347&0.336\end{tabular}\\
Average&\begin{tabular}{P{0.8cm}|P{0.9cm}|c}\textbf{0.508}&0.317&0.320\end{tabular}&\begin{tabular}{P{0.8cm}|P{0.9cm}|c}\textbf{0.432}&0.293&0.338\end{tabular}&\begin{tabular}{P{0.8cm}|P{0.9cm}|c}\textbf{0.924}&0.862&0.860\end{tabular}&\begin{tabular}{P{0.8cm}|P{0.9cm}|c}\textbf{0.759}&0.512&0.433\end{tabular}\\
\end{tabular}
\caption{Quantitative comparison on sequences in the test set of Ego2HandsPose.}
\label{tab:table_seq_comparison}
\end{table*}
\section{Additional Results}
\begin{figure*}[t]
  \centering
  \begin{subfigure}{.33\linewidth}
    \includegraphics[width=.5\linewidth]{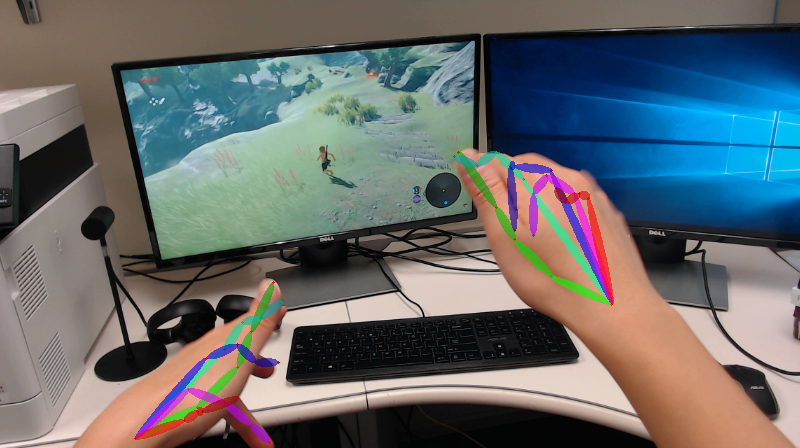}\hfill
    \includegraphics[width=.5\linewidth]{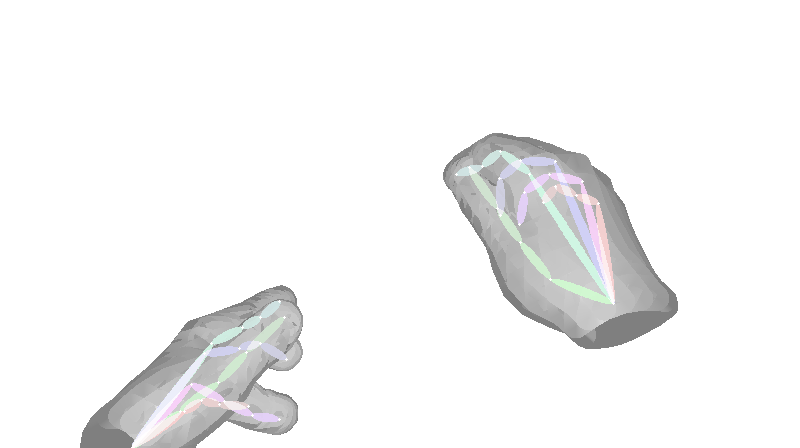}\hfill
  \end{subfigure}
  \begin{subfigure}{.33\linewidth}
    \includegraphics[width=.5\linewidth]{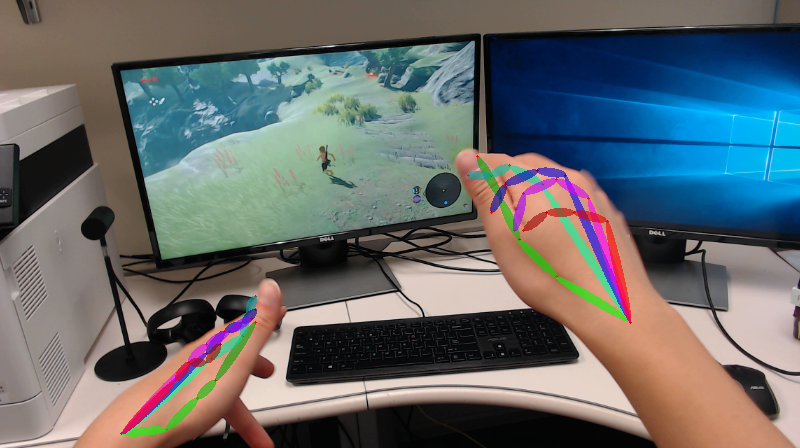}\hfill
    \includegraphics[width=.5\linewidth]{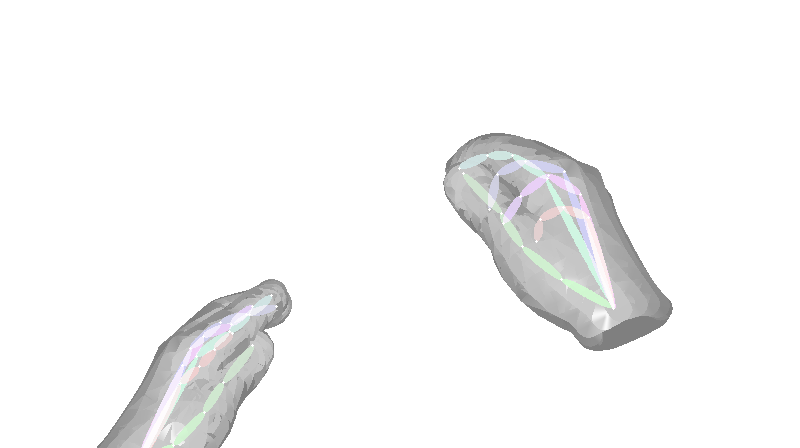}\hfill
  \end{subfigure}
  \begin{subfigure}{.33\linewidth}
    \includegraphics[width=.5\linewidth]{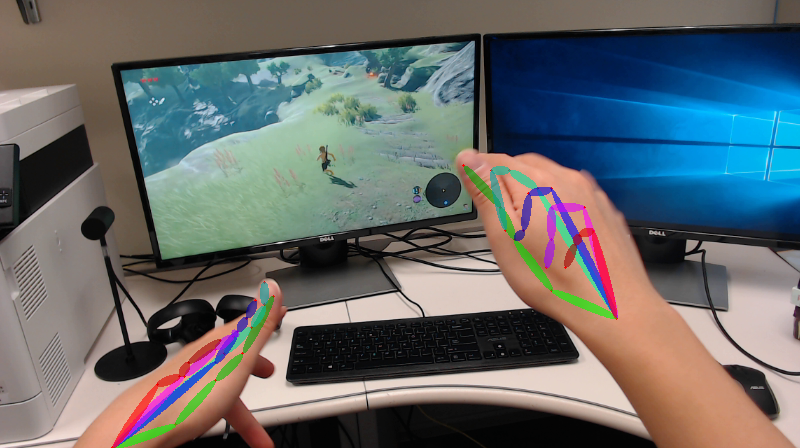}\hfill
    \includegraphics[width=.5\linewidth]{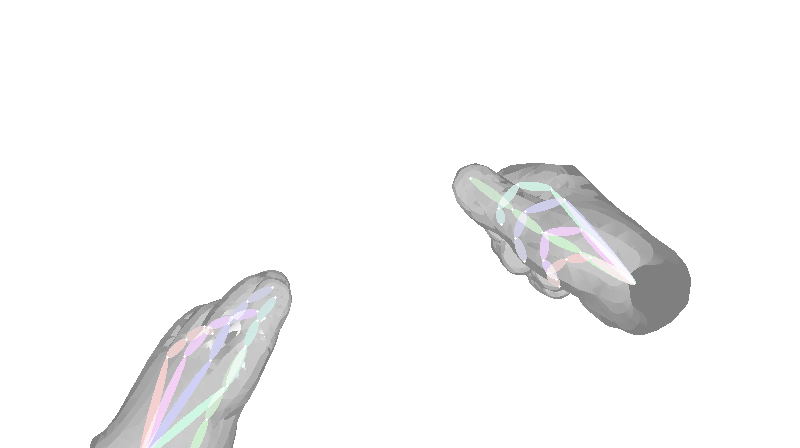}\hfill
  \end{subfigure}
  \begin{subfigure}{.33\linewidth}
    \includegraphics[width=.5\linewidth]{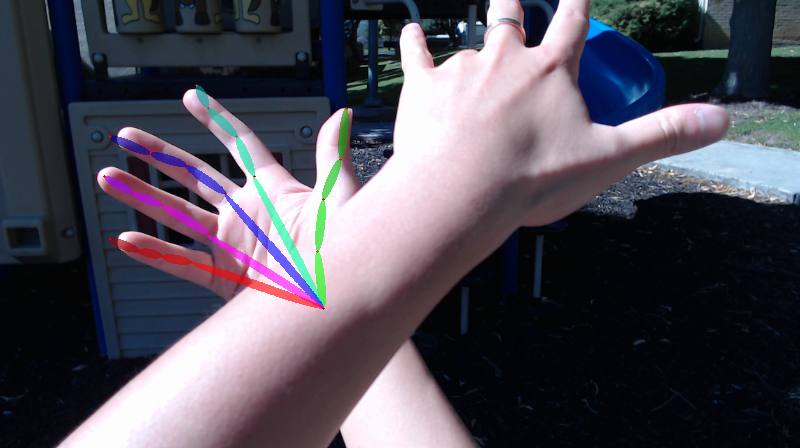}\hfill
    \includegraphics[width=.5\linewidth]{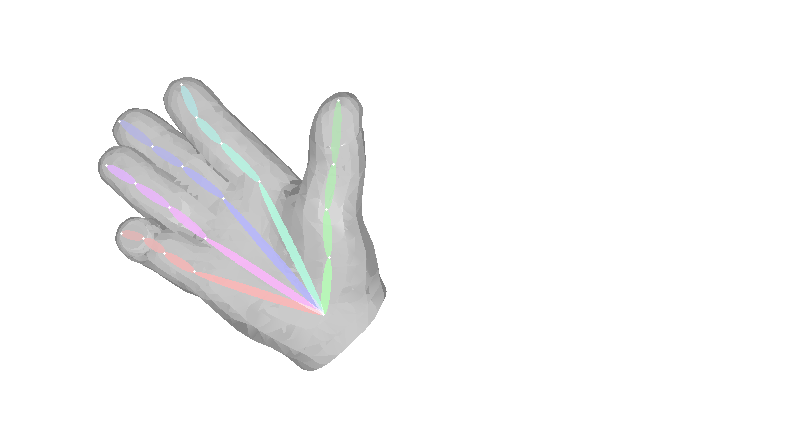}\hfill
  \end{subfigure}
  \begin{subfigure}{.33\linewidth}
    \includegraphics[width=.5\linewidth]{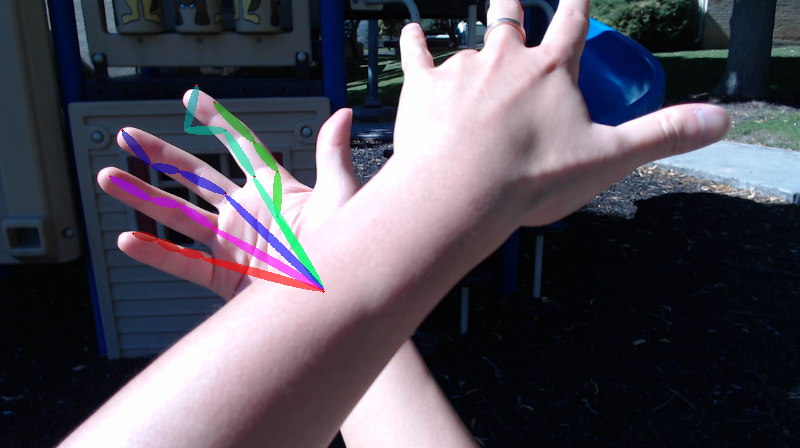}\hfill
    \includegraphics[width=.5\linewidth]{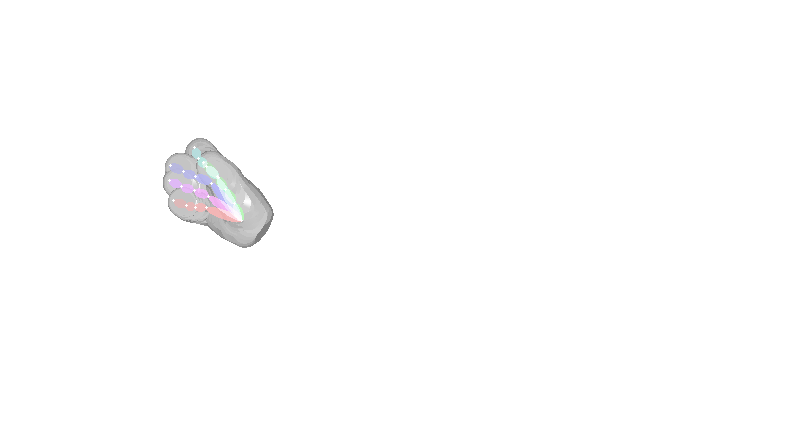}\hfill
  \end{subfigure}
  \begin{subfigure}{.33\linewidth}
    \includegraphics[width=.5\linewidth]{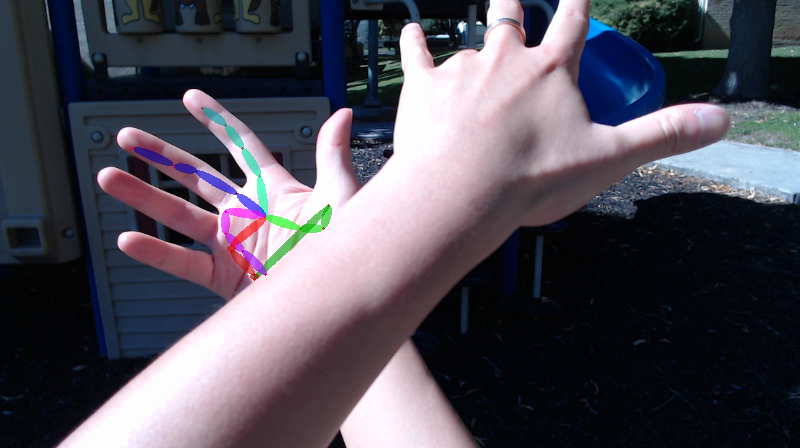}\hfill
    \includegraphics[width=.5\linewidth]{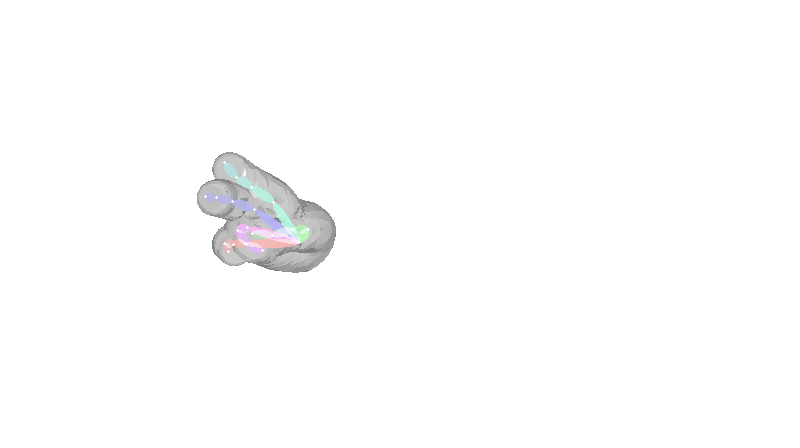}\hfill
  \end{subfigure}
  \begin{subfigure}{.33\linewidth}
    \includegraphics[width=.5\linewidth]{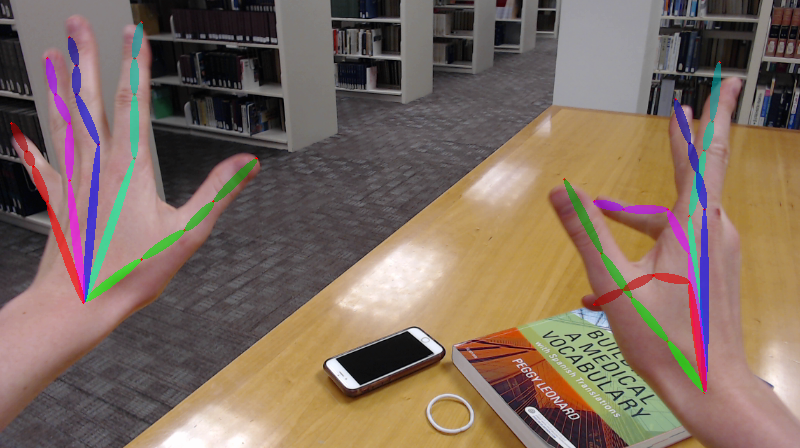}\hfill
    \includegraphics[width=.5\linewidth]{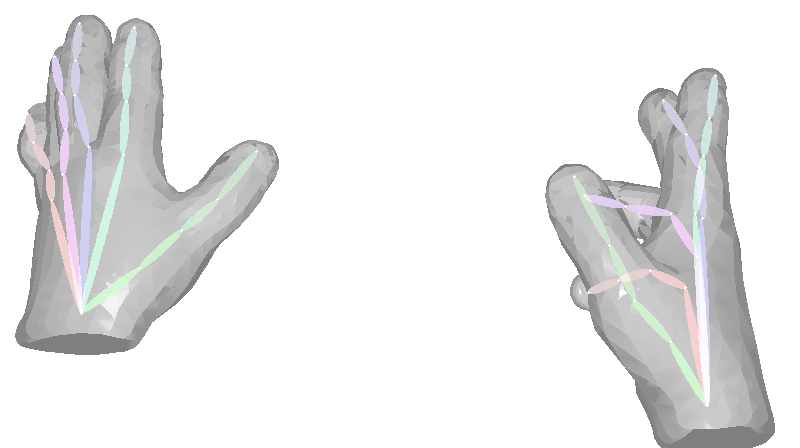}\hfill
  \end{subfigure}
  \begin{subfigure}{.33\linewidth}
    \includegraphics[width=.5\linewidth]{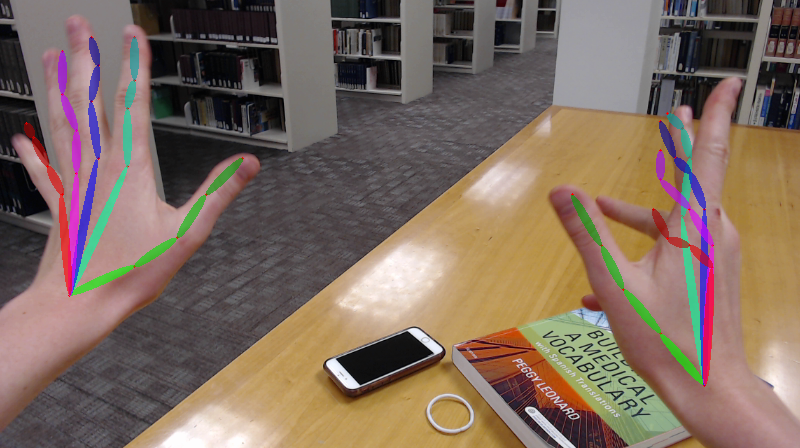}\hfill
    \includegraphics[width=.5\linewidth]{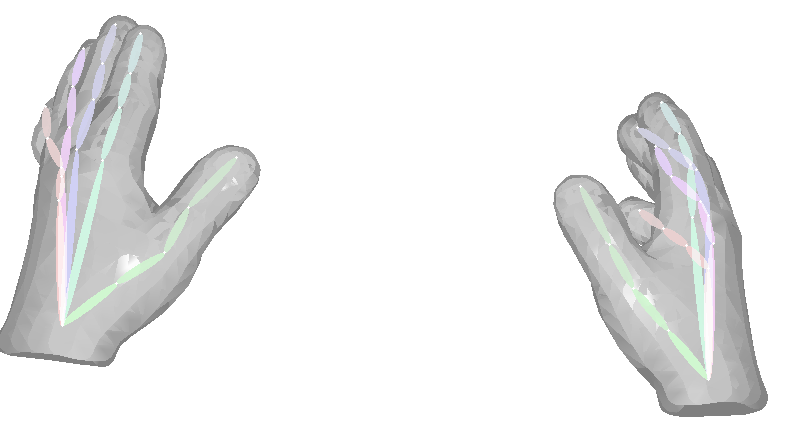}\hfill
  \end{subfigure}
  \begin{subfigure}{.33\linewidth}
    \includegraphics[width=.5\linewidth]{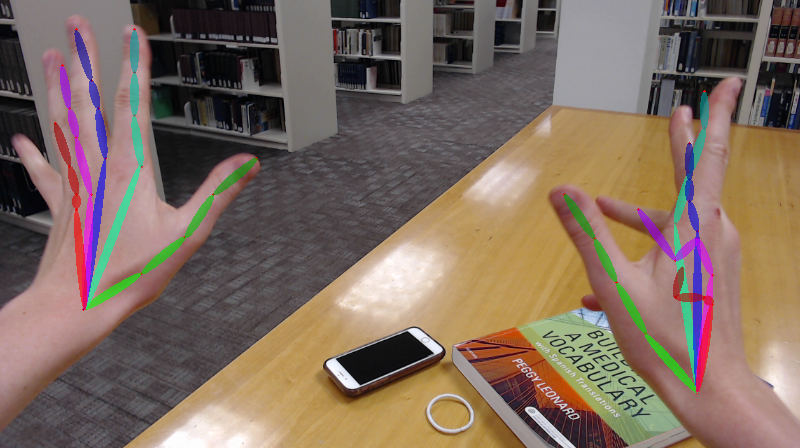}\hfill
    \includegraphics[width=.5\linewidth]{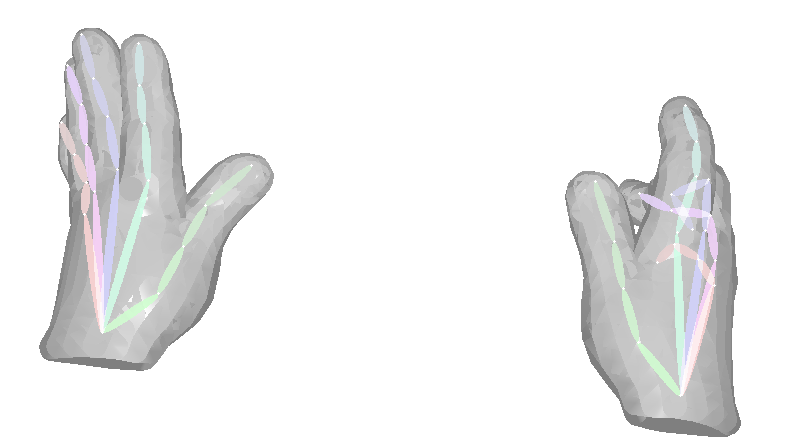}\hfill
  \end{subfigure}
  \begin{subfigure}{.33\linewidth}
    \includegraphics[width=.5\linewidth]{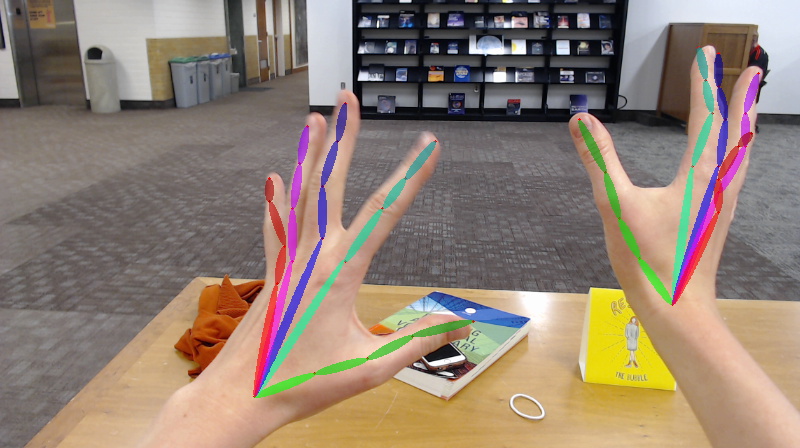}\hfill
    \includegraphics[width=.5\linewidth]{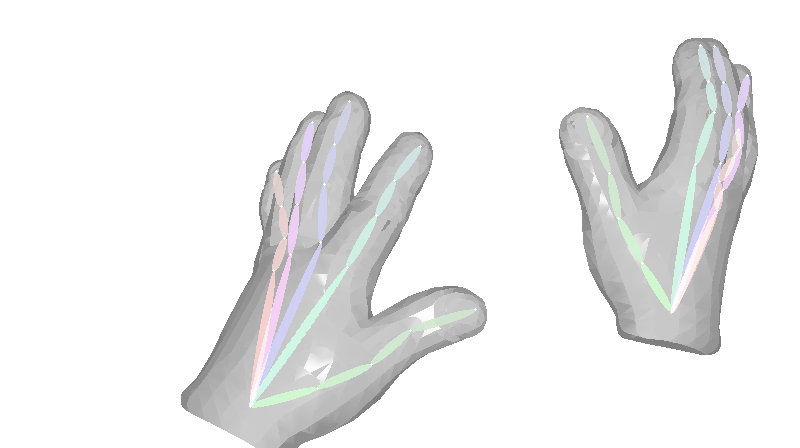}\hfill
  \end{subfigure}
  \begin{subfigure}{.33\linewidth}
    \includegraphics[width=.5\linewidth]{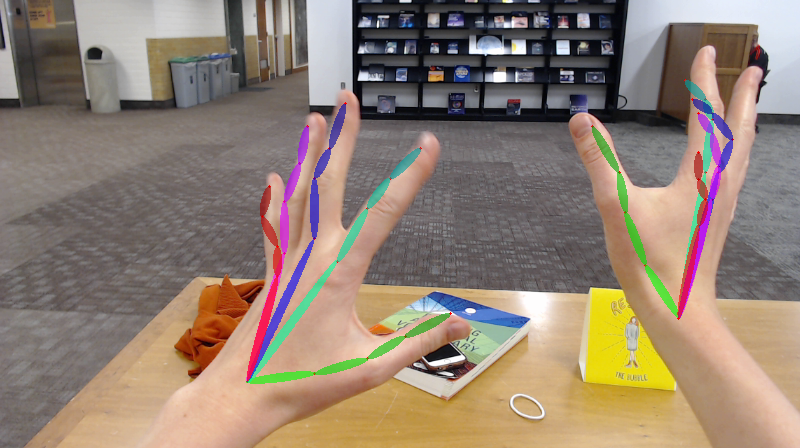}\hfill
    \includegraphics[width=.5\linewidth]{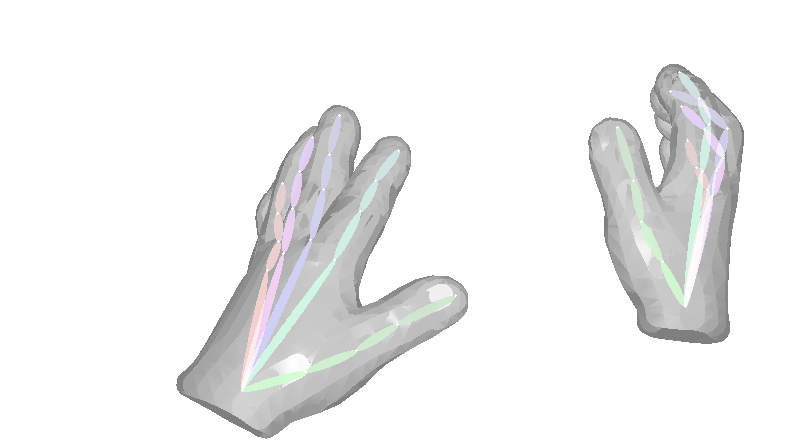}\hfill
  \end{subfigure}
  \begin{subfigure}{.33\linewidth}
    \includegraphics[width=.5\linewidth]{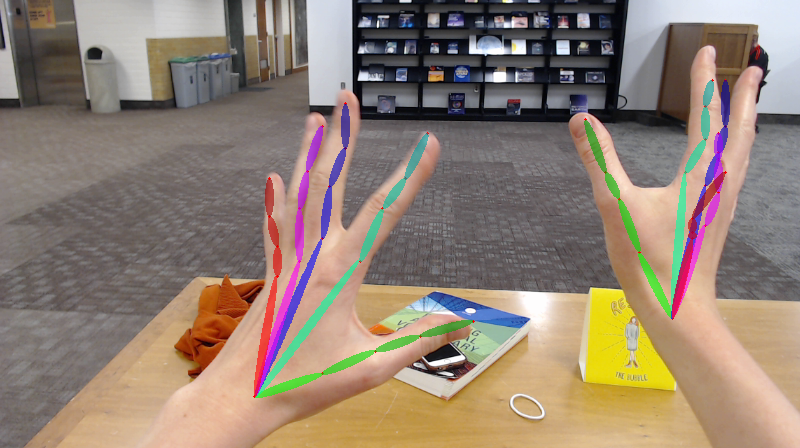}\hfill
    \includegraphics[width=.5\linewidth]{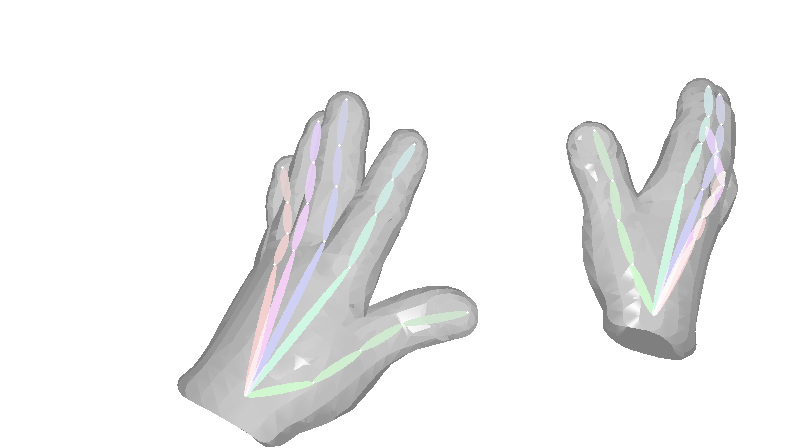}\hfill
  \end{subfigure}
  \begin{subfigure}{.33\linewidth}
    \includegraphics[width=.5\linewidth]{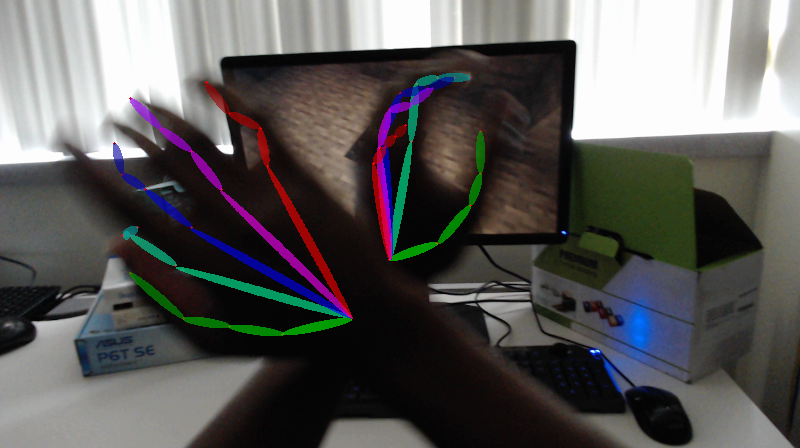}\hfill
    \includegraphics[width=.5\linewidth]{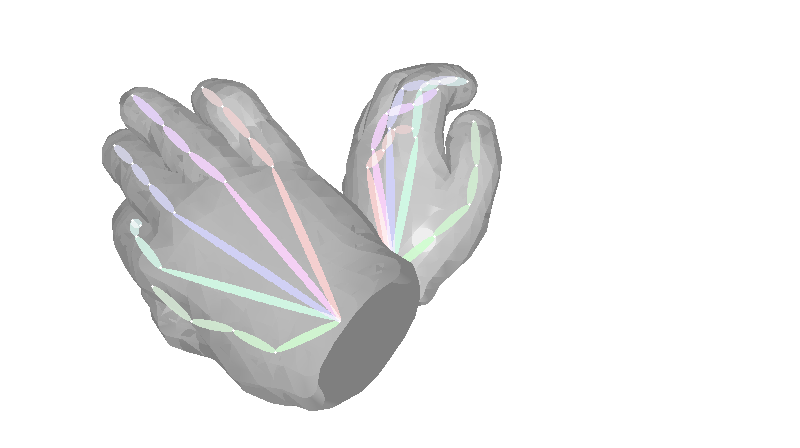}\hfill
  \end{subfigure}
  \begin{subfigure}{.33\linewidth}
    \includegraphics[width=.5\linewidth]{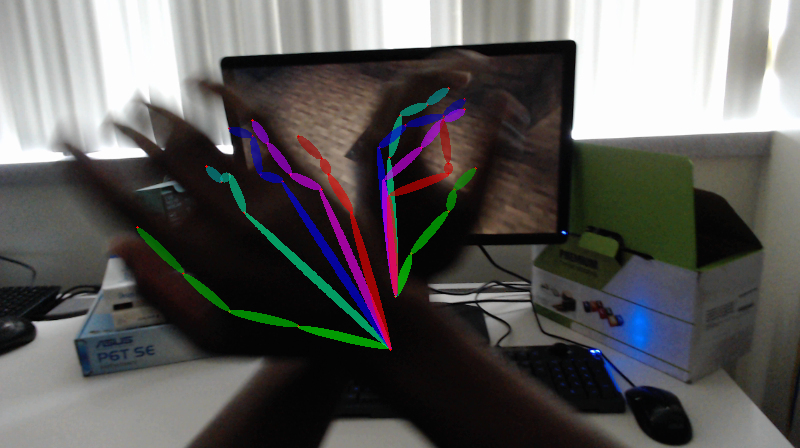}\hfill
    \includegraphics[width=.5\linewidth]{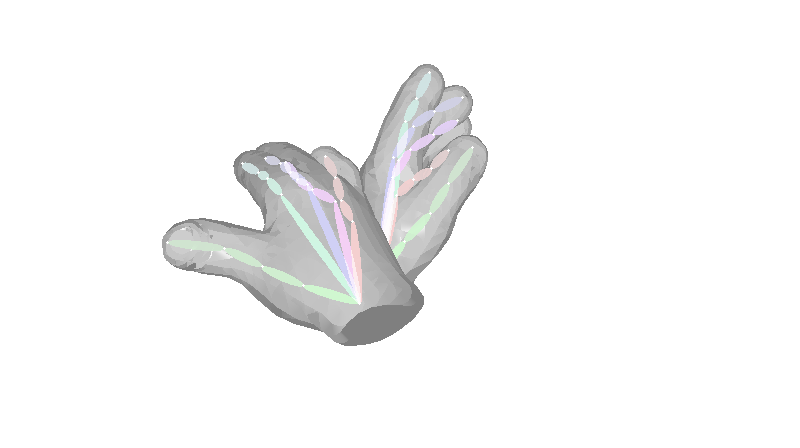}\hfill
  \end{subfigure}
  \begin{subfigure}{.33\linewidth}
    \includegraphics[width=.5\linewidth]{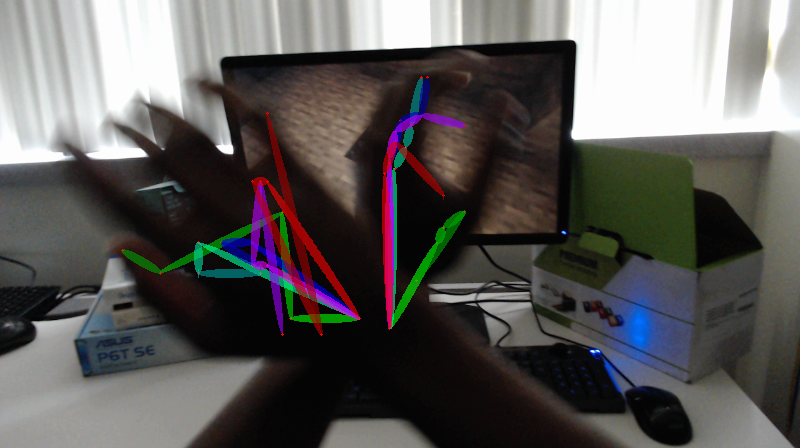}\hfill
    \includegraphics[width=.5\linewidth]{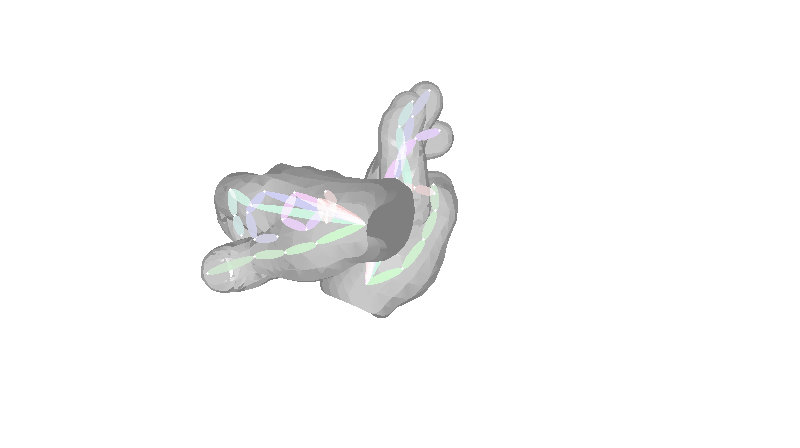}\hfill
  \end{subfigure}
  \begin{subfigure}{.33\linewidth}
    \includegraphics[width=.5\linewidth]{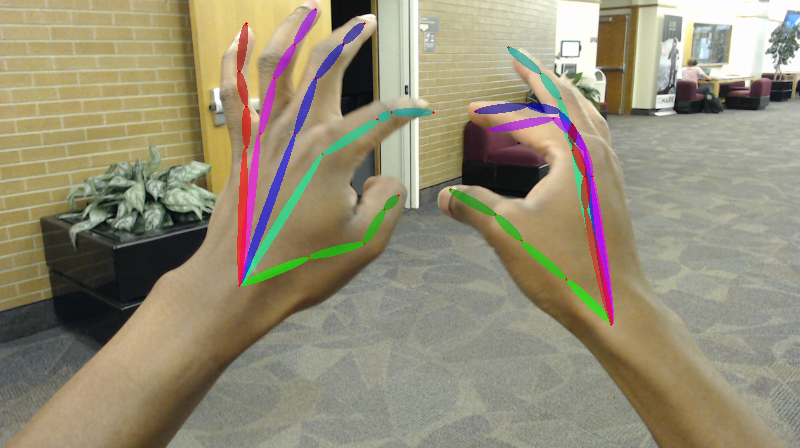}\hfill
    \includegraphics[width=.5\linewidth]{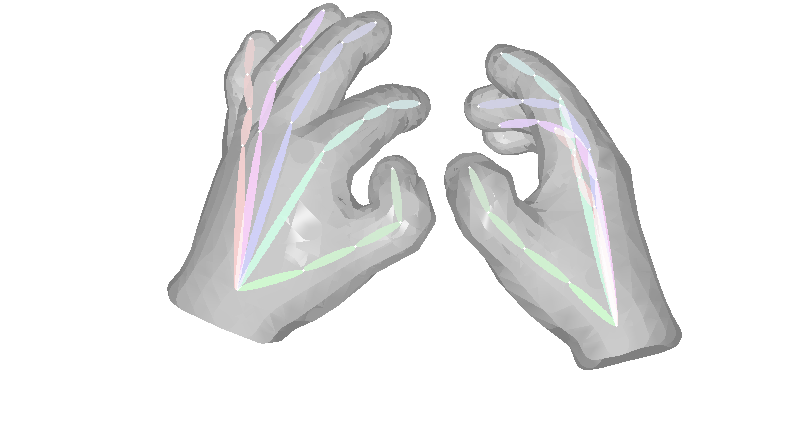}\hfill
  \end{subfigure}
  \begin{subfigure}{.33\linewidth}
    \includegraphics[width=.5\linewidth]{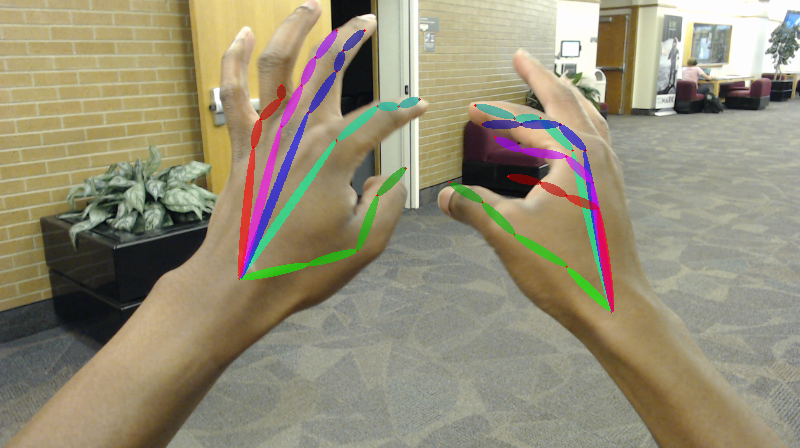}\hfill
    \includegraphics[width=.5\linewidth]{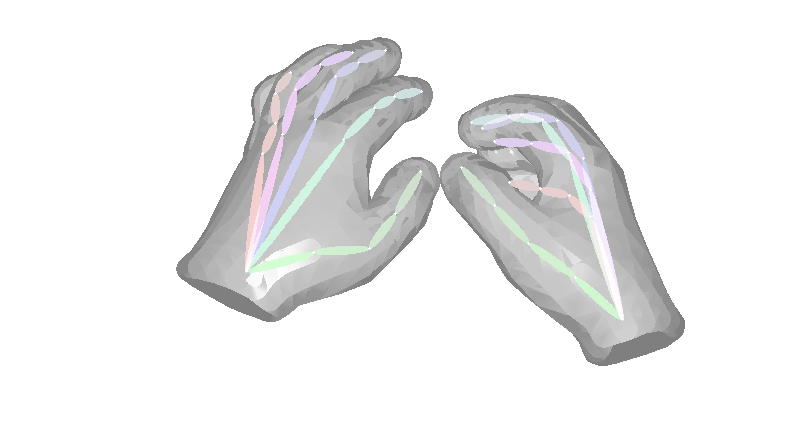}\hfill
  \end{subfigure}
  \begin{subfigure}{.33\linewidth}
    \includegraphics[width=.5\linewidth]{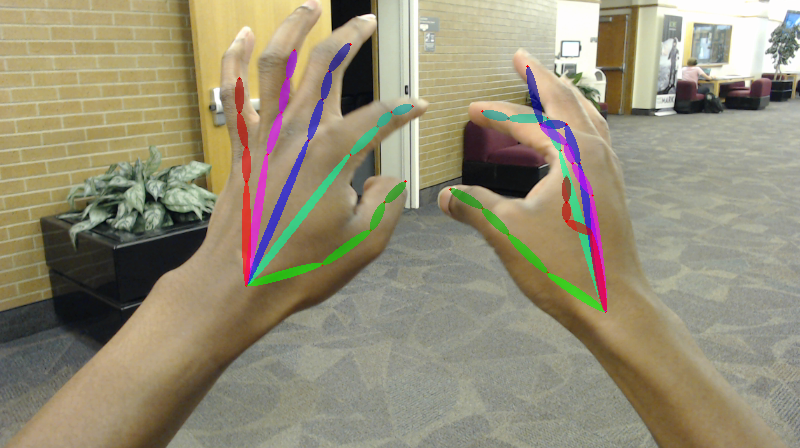}\hfill
    \includegraphics[width=.5\linewidth]{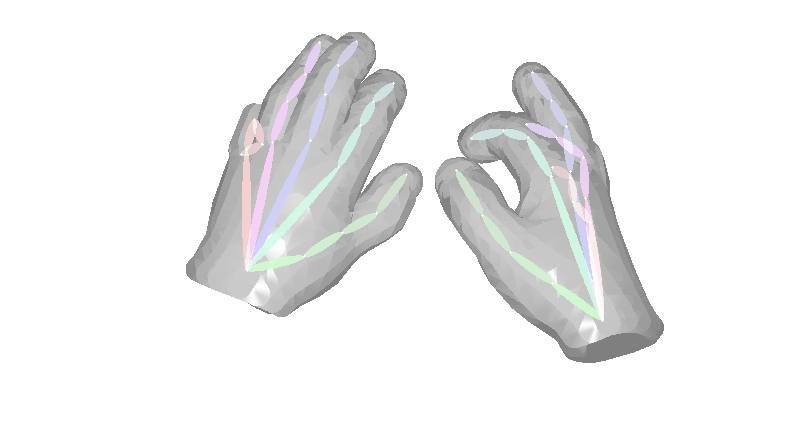}\hfill
  \end{subfigure}
  \begin{subfigure}{.33\linewidth}
    \includegraphics[width=.5\linewidth]{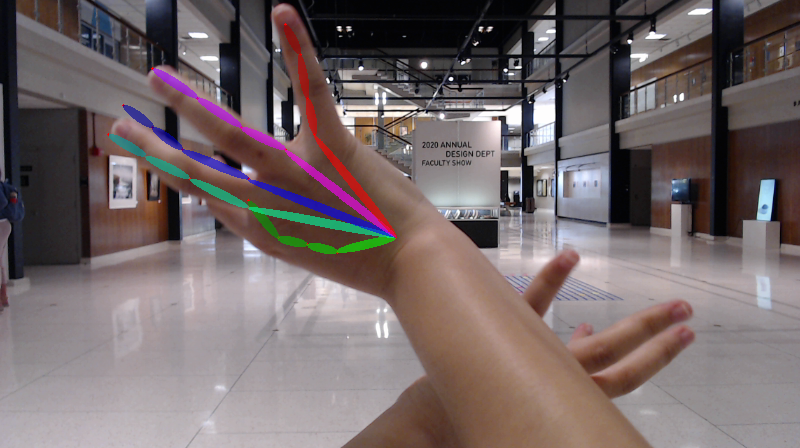}\hfill
    \includegraphics[width=.5\linewidth]{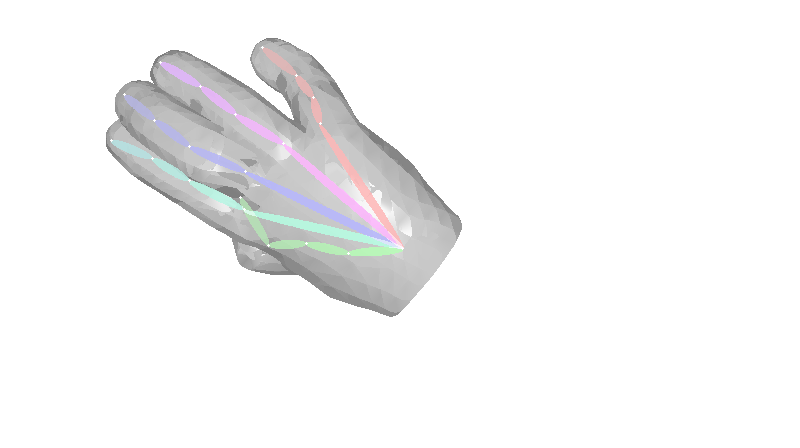}\hfill
  \end{subfigure}
  \begin{subfigure}{.33\linewidth}
    \includegraphics[width=.5\linewidth]{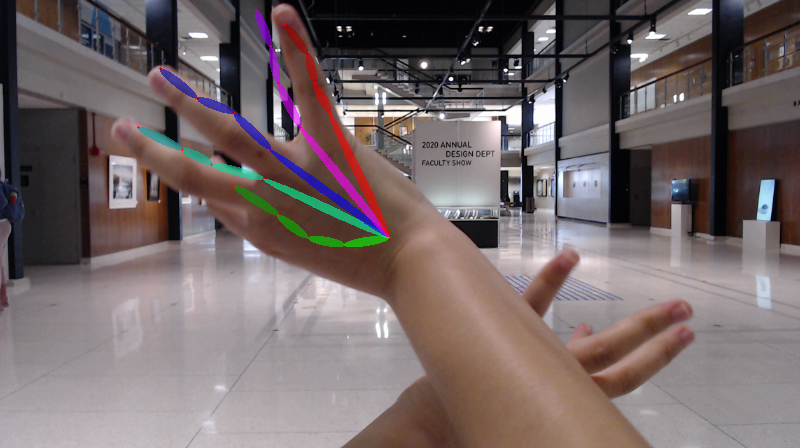}\hfill
    \includegraphics[width=.5\linewidth]{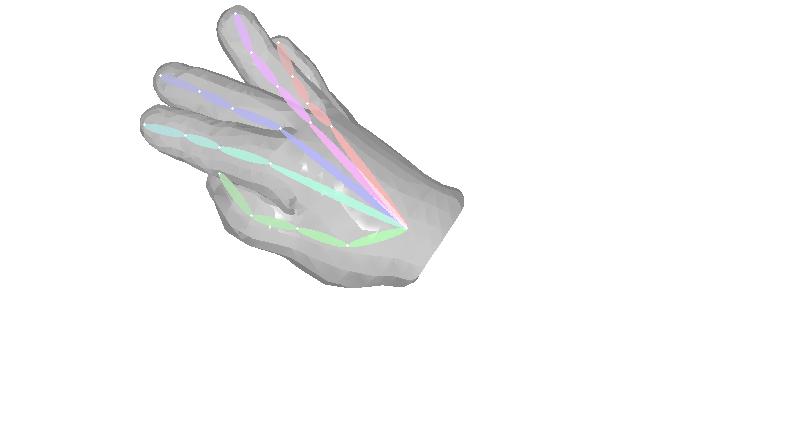}\hfill
  \end{subfigure}
  \begin{subfigure}{.33\linewidth}
    \includegraphics[width=.5\linewidth]{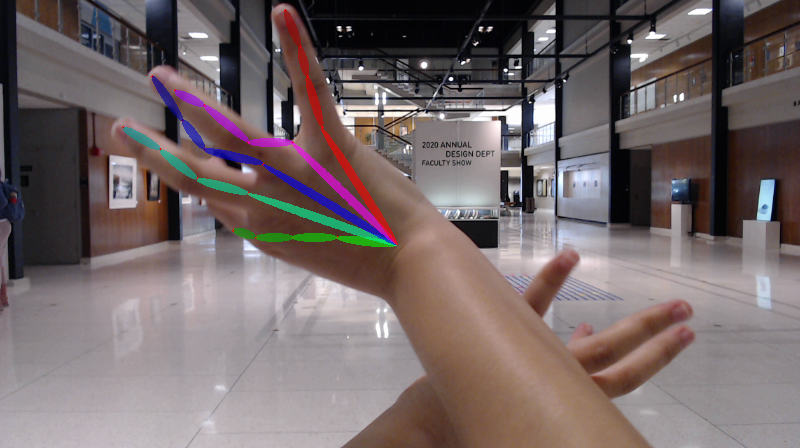}\hfill
    \includegraphics[width=.5\linewidth]{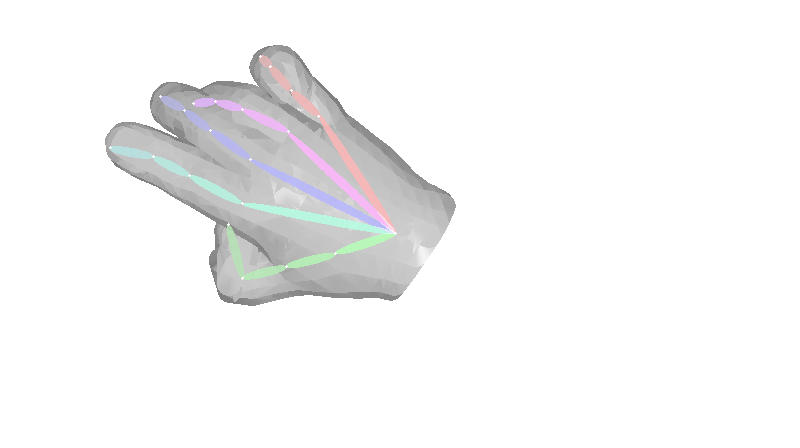}\hfill
  \end{subfigure}
  \begin{subfigure}{.33\linewidth}
    \includegraphics[width=.5\linewidth]{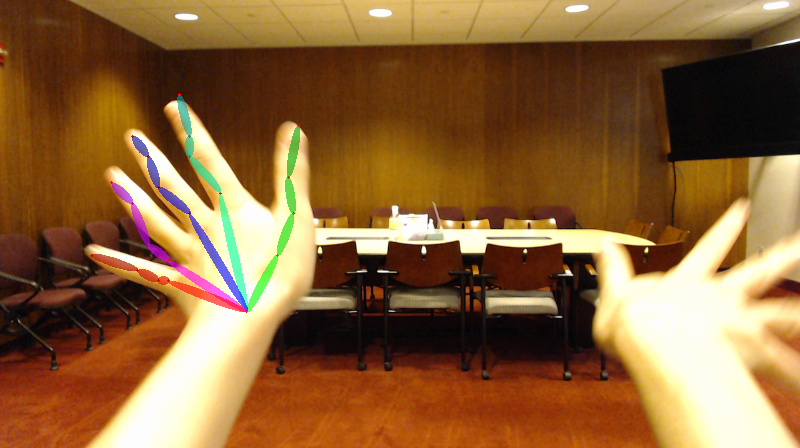}\hfill
    \includegraphics[width=.5\linewidth]{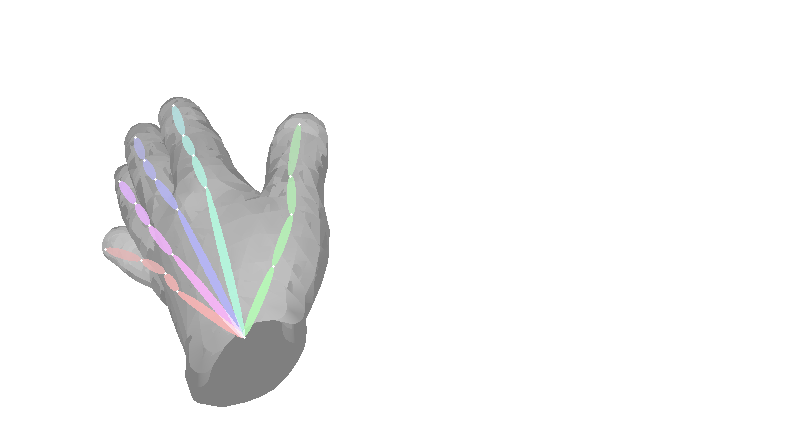}\hfill
    \caption{Ego2HandsPose}
  \end{subfigure}
  \begin{subfigure}{.33\linewidth}
    \includegraphics[width=.5\linewidth]{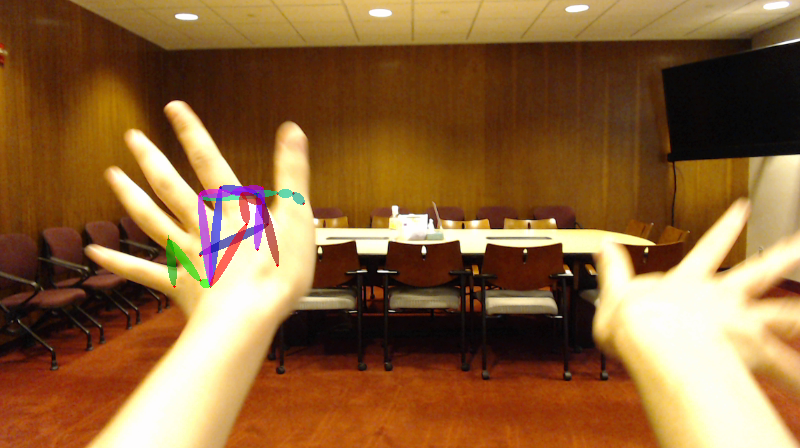}\hfill
    \includegraphics[width=.5\linewidth]{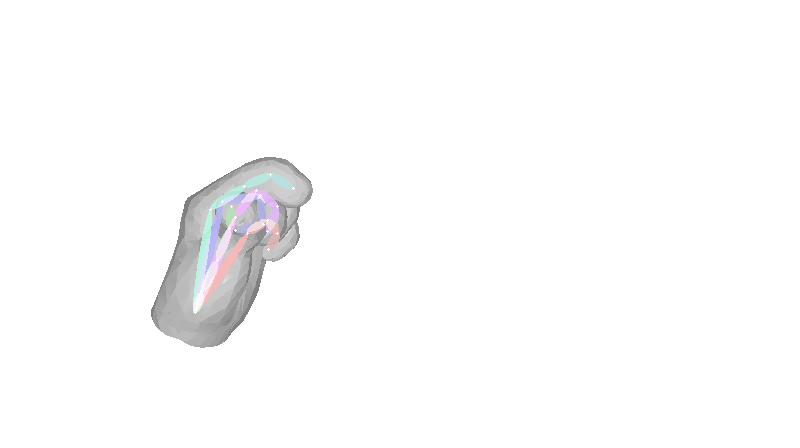}\hfill
    \caption{H2O}
  \end{subfigure}
  \begin{subfigure}{.33\linewidth}
    \includegraphics[width=.5\linewidth]{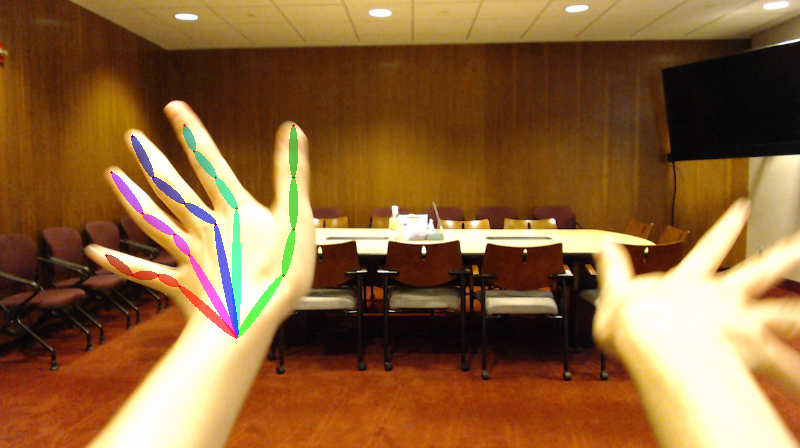}\hfill
    \includegraphics[width=.5\linewidth]{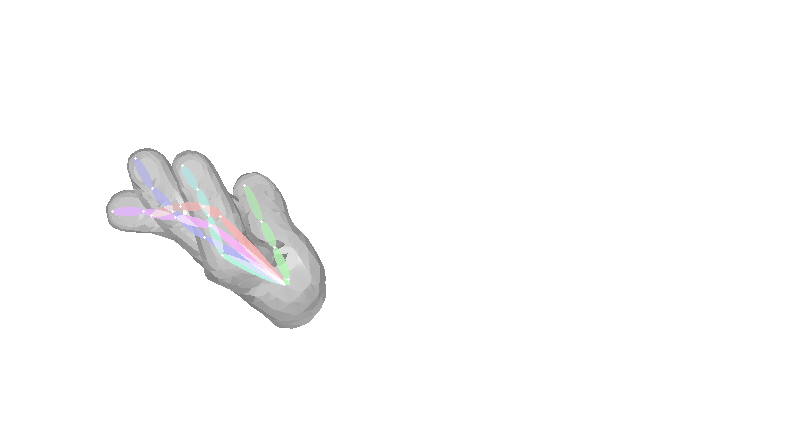}\hfill
    \caption{FreiHAND}
  \end{subfigure}
  \vspace{4mm}
  \caption{Qualitative comparison between models trained on selected datasets on sequences of Ego2HandsPose. Results from each dataset are shown in two columns. The first column displays input images with visualized estimated 2D hand poses. The second column displays the final global two-hand poses. Hand poses that are partially outside of the view are not estimated.}
  \label{fig:qualitative_comparison}
\end{figure*}

We provide quantitative results for all sequences in the test set of Ego2HandsPose in Table \ref{tab:table_seq_comparison}. Training on the training set of Ego2HandsPose outperforms other datasets on all sequences for all stages. We find that models trained on H2O and FreiHAND struggle particularly on sequences with bright (seq-2) and dark illumination (seq-5). In addition, these methods generally have lower performance on sequences with fast motion (seq-7 and seq-8), which introduces lower-quality input images challenging for 2D hand pose estimation. Qualitative comparison is provided in Figure \ref{fig:qualitative_comparison}.

\end{document}